\documentclass[lettersize,journal]{IEEEtran}

\bstctlcite{bstctl:etal, bstctl:nodash, bstctl:simpurl}
 
\usepackage{times}
\usepackage{epsfig}
\usepackage{graphicx}
\usepackage{amsmath,amsfonts,amssymb}
\usepackage{adjustbox}
\usepackage{subcaption}
\usepackage{enumitem}
\usepackage{textcomp}
\usepackage{multirow}
\usepackage{gensymb}
\usepackage{flushend}
\usepackage{xcolor}
\usepackage{colortbl}
\usepackage{hhline}
\usepackage{verbatim}
\usepackage{booktabs}
\usepackage{algorithm}
\usepackage{algorithmic}
\usepackage[colorlinks,pagebackref=false,citecolor=blue,bookmarks=false,hypertexnames=true]{hyperref}

\begin{document}

\title{Near-field Perception for Low-Speed Vehicle\\Automation using Surround-view Fisheye Cameras}
\author{Ciar\'{a}n~Eising$^{1\dag}$, Jonathan~Horgan$^{2\dag}$, and Senthil~Yogamani$^{2\dag}$\\
$^1$Department of Electronic and Computer Engineering, University of Limerick, Ireland \\
$^2$Valeo Vision Systems, Tuam, County Galway, Ireland \quad $^\dag$co-first authors
}

\maketitle


\textbf{\begin{abstract}
Cameras are the primary sensor in automated driving systems. They provide high information density and are optimal for detecting road infrastructure cues laid out for human vision. Surround-view camera systems typically comprise of four fisheye cameras with 190°+ field of view covering the entire 360° around the vehicle focused on near-field sensing. They are the principal sensors for low-speed, high accuracy, and close-range sensing applications, such as automated parking, traffic jam assistance, and low-speed emergency braking. In this work, we provide a detailed survey of such vision systems, setting up the survey in the context of an architecture that can be decomposed into four modular components namely Recognition, Reconstruction, Relocalization, and Reorganization. We jointly call this the \textit{4R Architecture}. We discuss how each component accomplishes a specific aspect and provide a positional argument that they can be synergized to form a complete perception system for low-speed automation. We support this argument by presenting results from previous works and by presenting architecture proposals for such a system. Qualitative results are presented in the video at \url{https://youtu.be/ae8bCOF77uY}.
\end{abstract}}

\section{Introduction} \label{sec:intro}

\begin{figure}[tb]
    \centering
    \includegraphics[width=\linewidth]{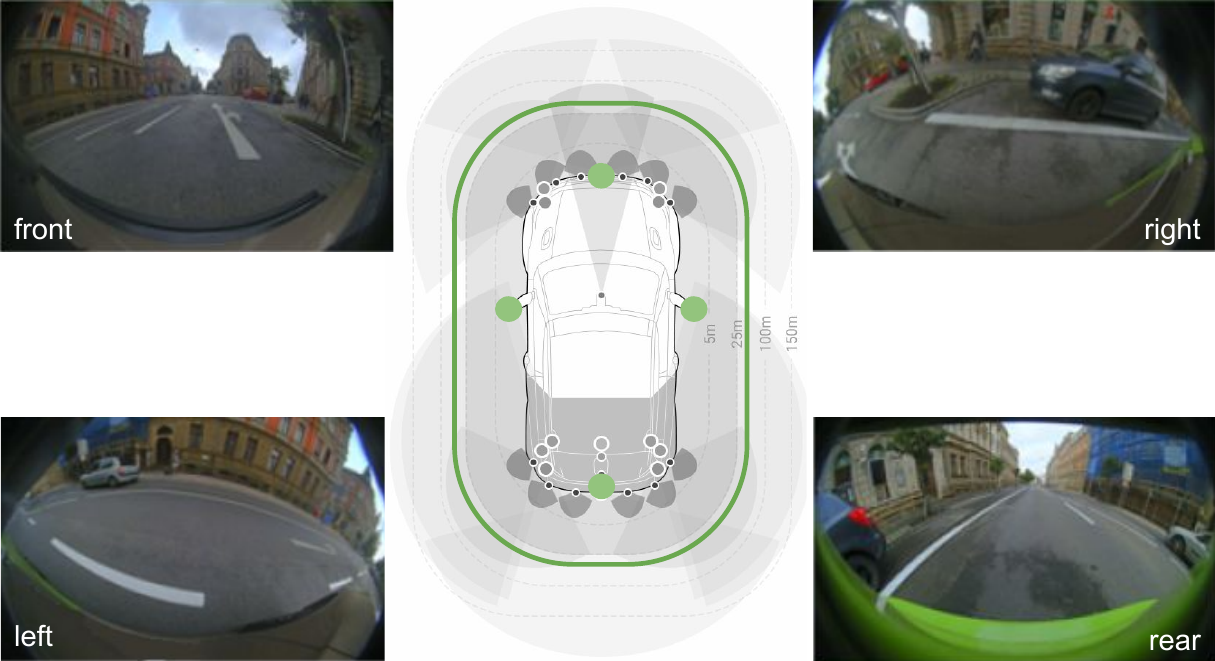}
     \caption{Images from the surround-view camera network. Green perimeter shows 360$^\circ$ near-field sensing around the vehicle.}
    \label{fig:svs}
\end{figure}

Recently, Autonomous Driving (AD) gained huge attention with significant progress in deep learning and computer vision algorithms \cite{9046805}. Within the next 5-10 years, AD is expected to be deployed commercially \cite{Stricker2020}, with widespread deployment in the coming decades. Currently, most automotive original equipment manufacturers (OEMs) are working on development projects focusing on autonomous driving technology \cite{cb2020commercial}, with computer vision having high importance \cite{badue2020autonomy}. However, as more is asked from computer vision systems deployed for vehicle autonomy, the architectures of such systems become ever more complex. Thus, it is of advantage to take a step back, and consider the architectures at the highest level. While what we propose should be considered a general discussion on the structure of automotive computer vision systems, we will use specific examples of computer vision applied to Fisheye camera networks, such as surround-view/visual cocoon (see Figure \ref{fig:svs}). 
In this paper, we aim to provide two elements to the reader. Firstly, we provide a comprehensive survey of automotive vision, with specific focus on near-field perception for low-speed maneuvering. Secondly, we make a positional argument that considering perception system architectures as specializations of the 4Rs of automotive computer vision leads to significant synergies and efficiencies.

\begin{figure}[tb]
    \centering
    \begin{subfigure}[b]{0.15\textwidth}
         \centering
         \includegraphics[trim={0.7cm 0.3cm 12.7cm 4.2cm},clip,width=\linewidth]{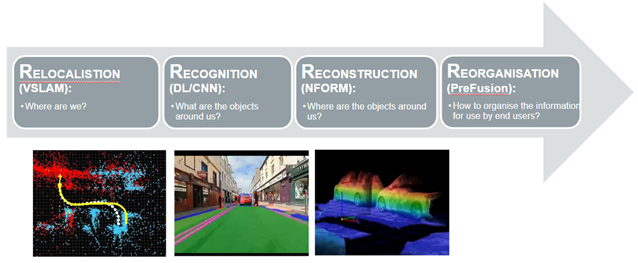}
         \caption{}
     \end{subfigure}
     \begin{subfigure}[b]{0.15\textwidth}
         \centering
         \includegraphics[trim={4.7cm 0.3cm 8.7cm 4.2cm},clip,width=\linewidth]{images/4R.png}
         \caption{}
     \end{subfigure}
     \begin{subfigure}[b]{0.15\textwidth}
         \centering
         \includegraphics[trim={8.7cm 0.3cm 4.7cm 4.2cm},clip,width=\linewidth]{images/4R.png}
         \caption{}
     \end{subfigure}
    \caption{Examples of (a) \textit{Relocalization}, (b) \textit{Recognition} and (c) \textit{Reconstruction} in action on a surround-view camera system.}
    \label{fig:4Rexamples}
    \vspace{-0.5cm}
\end{figure}

Several related surveys exist in the literature. Most notably, the work of Malik et al. \cite{malik2016_3rs}, which we shall discuss in more detail soon. A prior survey from the authors \cite{heimberger2017computer} discussed in detail the role of computer vision in automated parking applications. However, vision components were discussed mostly in isolation, with interdependencies between them not being discussed. In this work, while we briefly mention some embedded considerations, it is only in the context of architecture design. A more detailed overview of embedded considerations for driver assistance systems is provided in \cite{Velez2016}. Loce et al. \cite{Loce2013} provide a very high-level overview of vision systems for automotive, including traffic management, driver monitoring and security and law enforcement. Their treatment of on-vehicle vision systems is limited to lane keeping, pedestrian detection and driver monitoring. In \cite{Hughes2009WideangleCT}, one of the few surveys that discusses automotive surround-view systems is provided, though the focus is entirely on blind-zone viewing. The output of any perception system used in autonomous driving will be passed to the decision making, trajectory planning, and vehicle control system(s). While in this paper we focus solely on the visual perception elements, the reader is referred to \cite{BADUE2021113816} for a useful overview of autonomous driving systems.

The work of this paper is inspired, in part, by the work of Malik et al. in \cite{malik2016_3rs}. The authors of that work propose that the core problems of computer vision are reconstruction, recognition, and reorganization, what they dub as the 3Rs of Computer Vision. Here, we propose to extend and specialize the 3Rs of Computer Vision to the 4Rs of Automotive Computer Vision: \textit{Reconstruction}, \textit{Recognition}, \textit{Reorganization} and \textit{Relocalization}. Figure \ref{fig:4Rexamples} shows examples of the first three Rs.

As with \cite{malik2016_3rs}, {\em Reconstruction} means inferring scene geometry from a video sequence, including the position of the vehicle within the scene. The importance of this should be obvious, as it is central to problems in scene mapping, obstacle avoidance, maneuvering and vehicle control. Malik et al. extend this beyond just geometric inference to include properties such as reflectance and illumination. However, these additional properties are not (currently, at least) significant in the context of automotive computer vision, and so we define \textit{Reconstruction} in the more traditional sense of meaning 3D geometry recovery.

{\em Recognition} is the term used for attaching semantic labels to aspects of a video image or scene. As in \cite{malik2016_3rs}, hierarchies are included in recognition. For example, a cyclist has a spatial hierarchy, as it can be divided into the subsets of bicycle and rider, and a vehicle category can have taxonomic subcategories of car, lorry, bicycle, etc. This can continue as far as is useful for an autonomous driving system. Lights can be categorized by the type (vehicle light, streetlights, stop lights, etc.), color (\textcolor{red}{red}, \textcolor{orange}{yellow}, \textcolor{teal}{green}), and their importance to the autonomous vehicle (need to respond, can ignore), which infers higher level reasoning of the system.

{\em Relocalization} is place recognition and metric localization of a vehicle relative to its surroundings. Relocalization can happen against a pre-recorded trajectory in the host vehicle, for example, for trained parking \cite{heimberger2017computer}, or against a map that is transferred from the infrastructure, for example, HD Maps \cite{li2017hdmap}. It is highly related to loop closure in SLAM \cite{carranza2013reloc}, though rather than consider just the problem of loop closure, we consider the broader problem of the localization of the vehicle against one or many pre-defined maps.

{\em Reorganization} is the approach of combining information from the previous three components of computer vision into a unified representation. In the work of Malik et al., reorganization is derived from the term ``perceptual organization", and roughly equates it with segmentation. Image segmentation approaches are now dominated by CNN-based semantic segmentation and instance segmentation and therefore fall into the domain of recognition. In this paper, we use the term to equate with ``late fusion", which is the manipulation, filtering and reorganization of inputs into a unified output. This is an important step in the context of vehicle automation, as a unified representation of the sensor outputs is required for vehicle control. This also admits the fusion of the outputs of multiple cameras at a late stage and can be a pre-filter for automotive sensor fusion \cite{gustafsson2009safety}.

\begin{figure}[tb]
    \centering
    \includegraphics[width=0.6\linewidth]{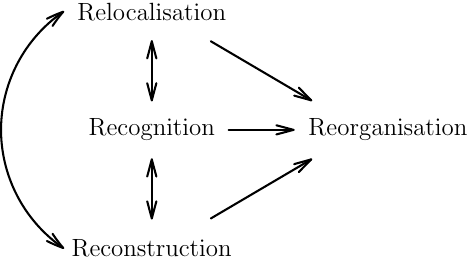}
    \caption{The 4Rs of automotive vision, and their information flow paths.}
\label{fig:4Rs_graph}
\end{figure}

The useful information flow paths are shown in Figure \ref{fig:4Rs_graph}. As we shall discuss, each of the three of Relocalization, Recognition and Reconstruction have useful information for the other, and all three feed the Reorganization component. It could be argued that even Reorganization has useful information for the other three components. While recurrent systems have seen some traction in specific automotive vision problems \cite{coello2019recurrent, borghi2017recurrent},
this is not commonplace at an architecture level in automotive visual systems.

Malik et al. \cite{malik2016_3rs} describe that in early years of computer vision, the concept of advantageous division of computer vision into low, medium, and high-level tasks was popular, but later became redundant. Later in this paper, we will argue that there is still merit to that mode of thinking, particularly when considering real-time embedded computer vision systems. For example, accessing the pixel information from an image is memory bandwidth intensive, and so doing all intensive pixel processing in a single step is logical. Therefore, we also propose a sub-division of computer vision into \textit{pipeline stages}, being \textit{Pixel Processing}, \textit{Intermediate Processing} and \textit{Object} stages, wrapped by \textit{pre-} and \textit{post-processing} stages. We will generally refer to this as the computer vision \textit{pipeline} or \textit{pipeline stages}.

Additionally, while what we discuss can, in the general sense, be applied to the entire field of automotive computer vision, we must acknowledge that convolutional neural networks (CNNs) have become the standard building block for many visual perception tasks in vehicle autonomy. Thus, many of our examples will focus on the role of neural networks. Bounding boxes for object detection is one of the first successful applications of CNNs for detecting not only pedestrians and vehicles, but also their positions. Recently semantic segmentation is becoming more mature \cite{siam2018rtseg, Qiao2021}, starting with detection of roadway composition like road surface, lanes, road markings, curbs, etc. CNNs are also becoming competitive for geometric vision tasks like depth estimation \cite{Kumar2018CVPRWorkshop} and Visual SLAM \cite{li2020slam}.

The rest of the paper is structured as follows. In Section \ref{sec:fwk}, we provide some background information on pertinent low-speed, near-field sensing use cases, we give an overview of fisheye cameras, and we provide a brief overview of the \textit{WoodScape} dataset \cite{yogamani2019woodscape}. This is the dataset we use to provide some results later in the paper. Section \ref{sec:arch} provides an overview of a surround-view sensing system architecture. Section \ref{sec:components} discusses the components of 4R in detail individually by providing a survey of work in each of the 4R areas. In Section \ref{sec:synergies}, we discuss the interactions between the 4R components, provide architecture proposals and give a set of results from prior works that support the architectural arguments.

\section{Overview of near-field sensing} \label{sec:fwk}

In this section, we will provide some background material that will give the reader a deeper understanding of what will come in the next sections. As most research in the autonomous driving perception field focuses on far-field perception using standard or narrow field of view cameras, it is pertinent to give some details here on near-field perception. We give an overview of near-field perception use cases, of how a fisheye camera differs from a standard camera, and of the WoodScape dataset that we use throughout the paper.

\subsection{Near-field sensing use cases}

Here we will discuss a few of the most pertinent use cases in vehicle autonomy for surround-view computer vision systems.

\subsubsection{Automated Parking Systems}
\begin{figure}[ht]
    \centering
    \includegraphics[width=\linewidth]{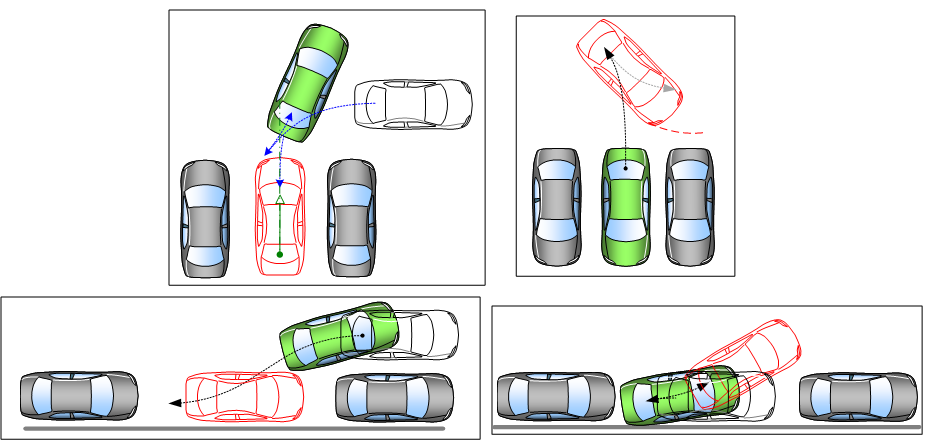}
     \caption{Typical Parking use cases. Top row: Perpendicular backward park in and perpendicular backward park out. Bottom row: Parallel backward parking in and park Out.}
\label{fig:parking_modes}
\end{figure}

\begin{figure}[ht]
     \centering
     \begin{subfigure}[b]{0.21\textwidth}
         \centering
         \includegraphics[width=\textwidth]{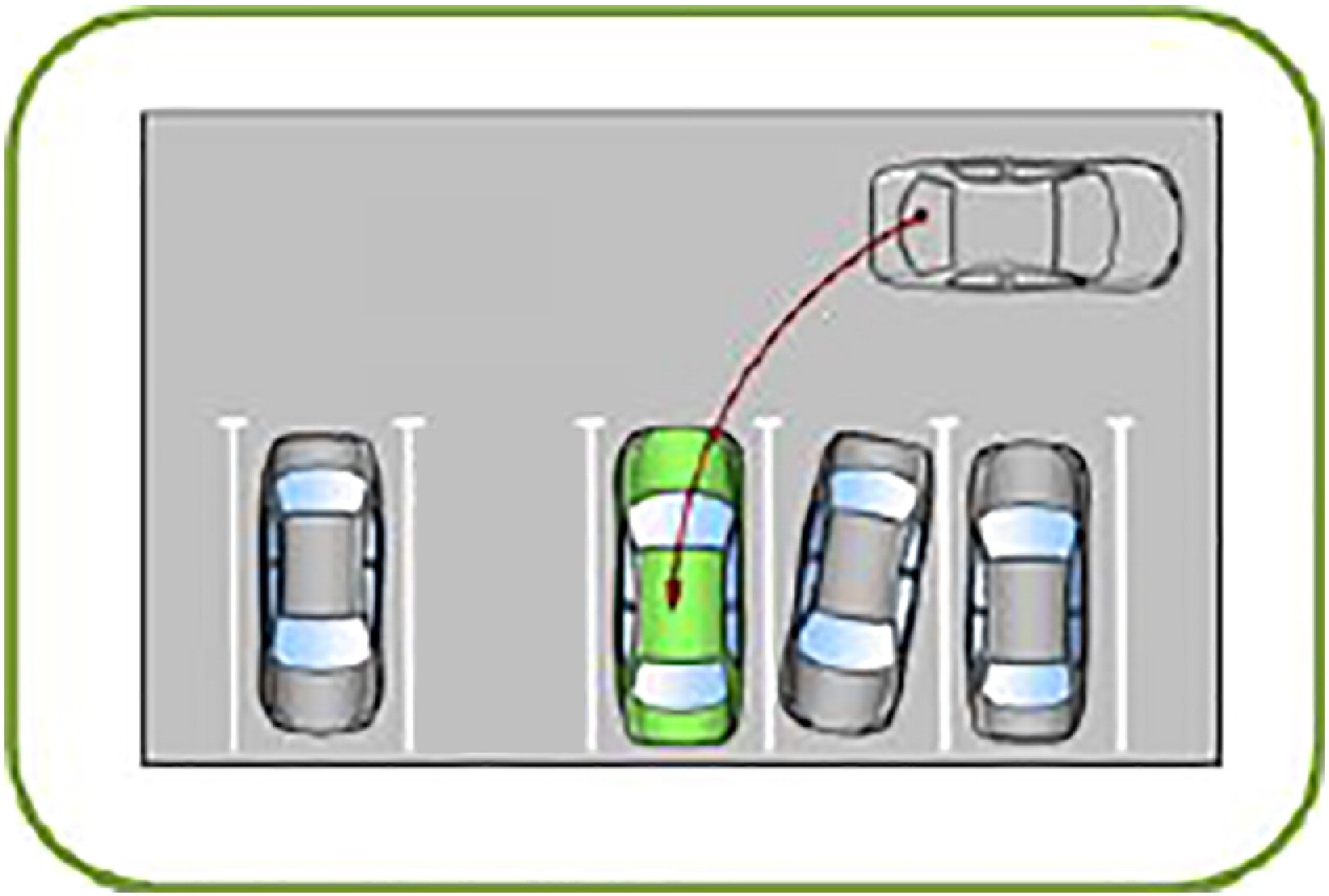}
     \end{subfigure}
     \begin{subfigure}[b]{0.24\textwidth}
         \centering
         \includegraphics[width=\textwidth]{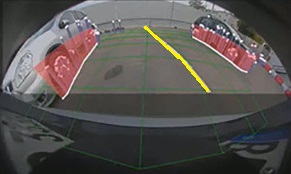}
     \end{subfigure}
        \caption{Parking with cameras can be guided by road markings and not just objects, unlike Ultrasonic and Radar. }
        \label{fig:camera_parking}
\end{figure}

Automated parking systems are one of the primary use cases for short range sensing \cite{heimberger2017computer}, with some typical parking use cases described in Figure \ref{fig:parking_modes}. As early as 1992, prototypes of semi-automated parking systems using radar systems were proposed, though not produced commercially \cite{walzer1990}. Early commercial partially automated parking systems employed either ultrasonic sensors or radar \cite{Wu2016, Song2016}. However, more recently, surround-view cameras are becoming one of the primary sensors for automated parking \cite{heimberger2017computer,Wang2014}. A major limitation of ultrasonic and radar sensors for automated parking is that parking slots can only be identified based on the presence of other obstacles (Figure \ref{fig:camera_parking}). Extending this, surround-view camera systems allow for parking in the presence of visual parking slot markings such as painted line markings, while also being seen as a key enabling technology for Valet Parking systems to become a reality \cite{Schwesinger2016, Ma2021}.


\subsubsection{Traffic Jam Assistance Systems}
\begin{figure}[ht]
    \centering
    \includegraphics[height=0.8\linewidth, angle=270]{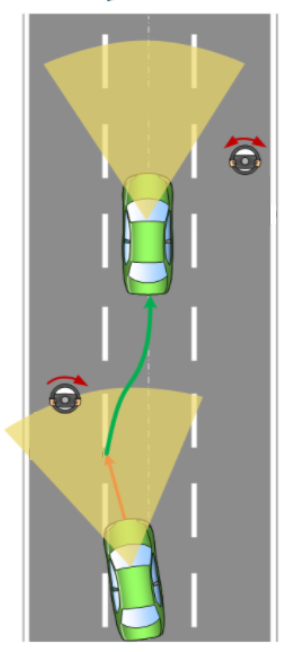}
     \caption{In traffic jam assist systems, the vehicle can assume both lateral and longitudinal control, including stopping and starting the motion of the vehicle.}
\label{fig:tja}
\end{figure}

\begin{figure}[ht]
     \centering
     \begin{subfigure}[b]{0.24\textwidth}
         \centering
         \includegraphics[width=\textwidth, trim=0 75 0 0, clip]{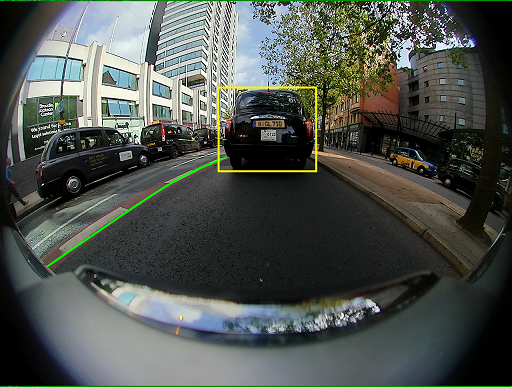}
     \end{subfigure}
     \begin{subfigure}[b]{0.24\textwidth}
         \centering
         \includegraphics[width=\textwidth, trim=0 75 0 0, clip]{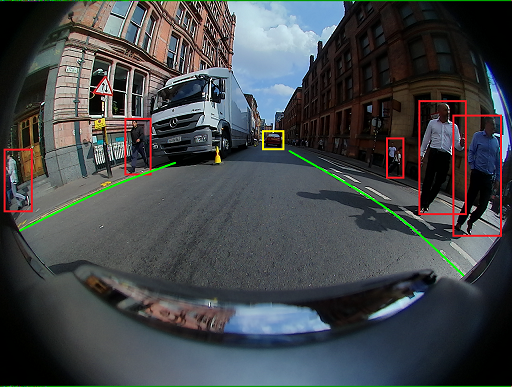}
     \end{subfigure}
        \caption{Surround-view cameras can be used for traffic jam assistance systems by detection leading vehicles and lane markings, for example. In addition, vulnerable road users (e.g., crossing pedestrians) that are outside the angular range of other sensors can be detected.}
        \label{fig:tja_sensing}
\end{figure}

As a substantial proportion of accidents are rear-end collisions at low speed \cite{Hellman2012}, traffic jam situations are considered to be one of the areas of driving that automation can give benefit in the short term \cite{Luke2014}, though current systems perhaps lack robustness \cite{Ulrich2019}. In automated traffic jam assistance systems, the vehicle assumes control of the longitudinal and lateral position while in the traffic jam scenario (Figure \ref{fig:tja}). This functionality is typically used in low speed environments, with maximum speeds of $\sim$60kph \cite{Ulrich2019}, though even lower maximum speeds of 40kph are suggested \cite{Rao2019}. While typically highway scenarios are considered for traffic jam assistance \cite{Luke2014}, there has been investigation into the urban traffic jam assistance systems \cite{Nothdurft2011}. Given the low-speed nature of this application, surround-view cameras are an ideal sensor, particularly in urban settings where, for example, pedestrians can attempt to cross from areas that are outside the field of view of traditional forward-facing cameras or radar systems. Figure \ref{fig:tja_sensing} shows examples of using surround-view cameras for traffic jam assist. In addition to detecting other road users and markings, features such a depth estimation \cite{Kumar2018CVPRWorkshop} and SLAM \cite{li2020slam} are also important for inferring distances to objects and controlling the vehicle position.


\subsubsection{Low Speed Braking}
Protection of vulnerable road users in low speed reversing situations has become a focus of legislation in some jurisdictions \cite{Sunstein2019}, with initial efforts to simply display the rearward portion of the vehicle to the driver \cite{Hughes2009WideangleCT}. It is shown in one study that automatic rearward braking significantly reduced collision claim rates \cite{CICCHINO201941}, with vehicles equipped with rear camera, parking assistance and automatic braking showing a 78\% reduction of reported collisions. Surround-view camera systems are extremely useful for low-speed braking, as the combinations of depth estimation and object detection are building blocks for this functionality.

\subsection{Fisheye cameras}

Fisheye cameras offer a distinct advantage for automotive applications. Given their extremely wide field of view, they can observe the full surrounding of a vehicle with a minimal number of sensors. 
Typically four cameras is all that is required for full 360$^\circ$ coverage of a car (Figure \ref{fig:svs}). 
However, this advantage comes with a cost given the significantly more complex projection geometry. Several papers in the past have provided reviews of how to model fisheye geometry, e.g., \cite{Usenko2018}. We do not aim to repeat this here and will rather focus on the problems that the use of fisheye camera technology brings to automotive visual perception.

\begin{figure}[tb]
    \centering
    \includegraphics[width=0.8\linewidth]{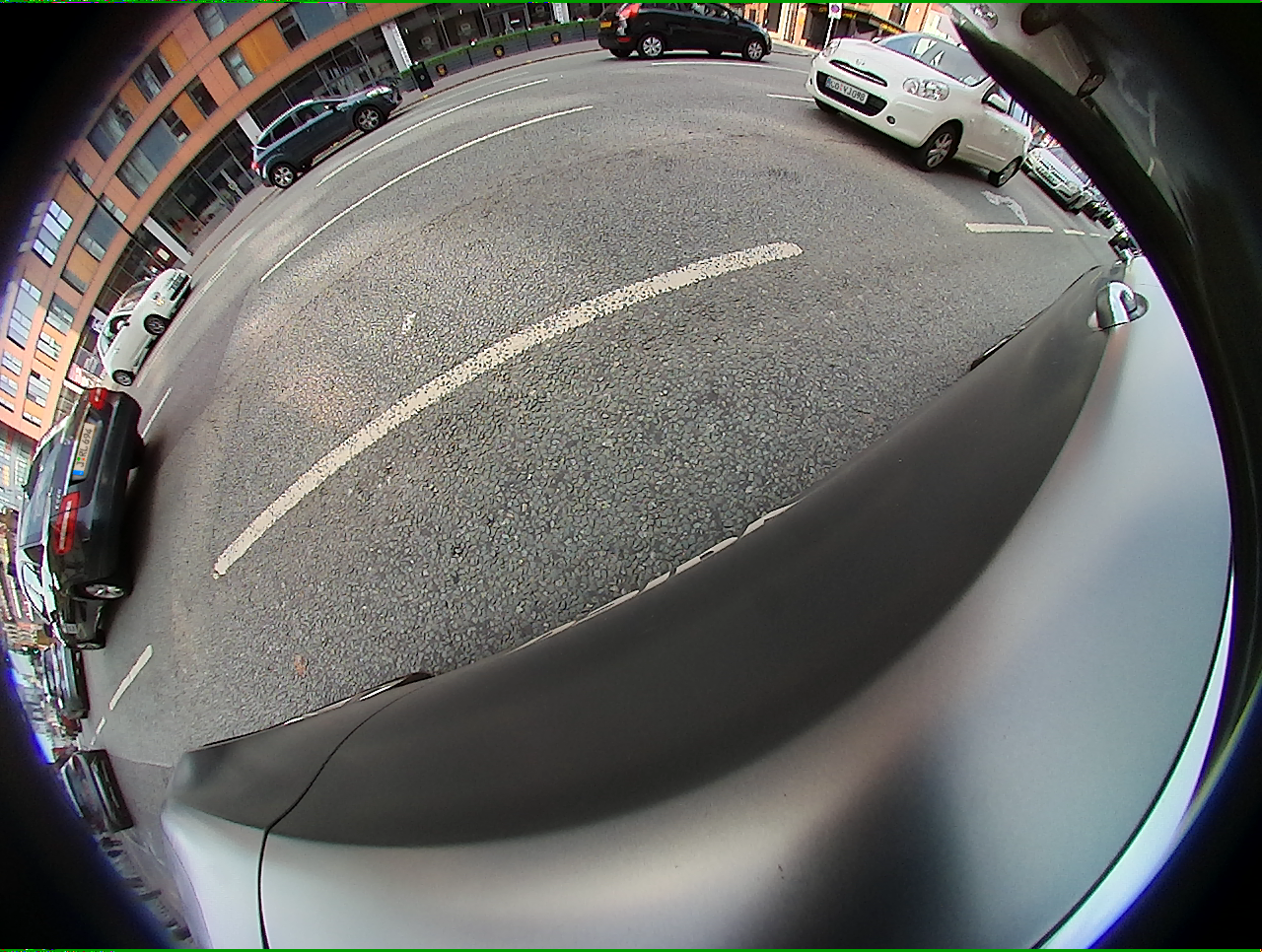}
    \caption{Significant deviation from translation invariance is evident in fisheye cameras.}
    \label{fig:mvl_camera}
\end{figure}

In standard field of view cameras, the principles of rectilinear projection and perspective are closely approximated, with the usual perspective properties, i.e., straight lines in the real world are projected as straight lines on the image plane. Parallel sets of straight lines are projected as a set of lines that are convergent on a single vanishing point on the image plane. Deviations from this through optical distortions are easily corrected. Many automotive datasets provide image data with optical distortions removed \cite{Geiger2013IJRR}, with an easy means for correction \cite{RobotCarDatasetIJRR}, or with almost imperceptible optical distortion \cite{cordts2016cityscapes}. As such, most research in automotive vision makes an implicit assumption of rectilinear projection. Fish-eye perspective differs significantly from rectilinear perspective. A straight line in the camera scene is projected as a curved line on the fish-eye image plane, and parallel sets of lines are projected as a set of curves that converge at two vanishing points \cite{HUGHES2010538}. This distortion is immediately obvious in the examples presented in this paper (e.g., Figures \ref{fig:svs}, \ref{fig:4Rexamples} and \ref{fig:camera_parking}). However, distortion is not the only effect. Figure \ref{fig:mvl_camera} shows an image from a typical mirror mounted camera in a surround-view system. In a fisheye camera, the orientation in the image of objects depends on their location in the image. In this example, the vehicle on the left is rotated almost 90$^\circ$ compared to the vehicle on the right. This has an impact on the translation invariance assumed in convolutional approaches to object detection. In standard cameras, translation invariance is an acceptable assumption. However, this is not the case in fisheye imagery, as evidenced in Figure \ref{fig:mvl_camera}. One must carefully consider how to handle this in any computer vision algorithm design.

\begin{figure}[tb]
    \centering
    \includegraphics[width=0.8\linewidth]{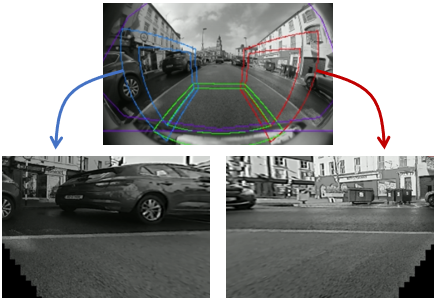}
    \caption{Warping on to multiple planes. Perspective distortion is evident in the left-hand image, and interpolation artefacts (blurring) is visible in both images.}
    \label{fig:multilinear}
\end{figure}

\begin{figure}[tb]
    \centering
    \includegraphics[width=\linewidth]{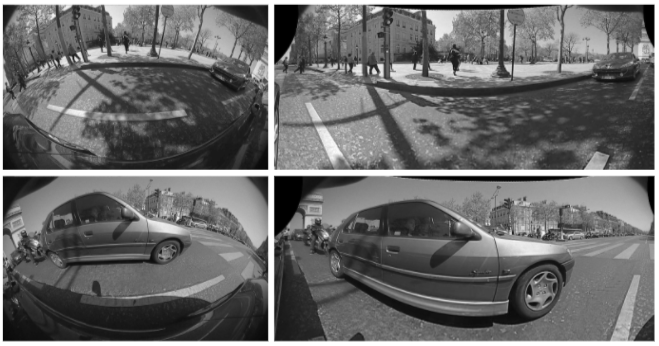}
    \caption{Warping to a cylindrical surface. In the automotive scene, vertical objects are vertical in the cylindrical image, and if they are not too close to the camera, appear like a perspective camera (observe pedestrians and cars in the top row). However, closer, and larger objects can be even more distorted in the cylindrical mapping.}
    \label{fig:cylindrical}
\end{figure}

\begin{figure}[ht]
     \centering
     \begin{subfigure}[b]{0.2\textwidth}
         \centering
         \includegraphics[width=\textwidth]{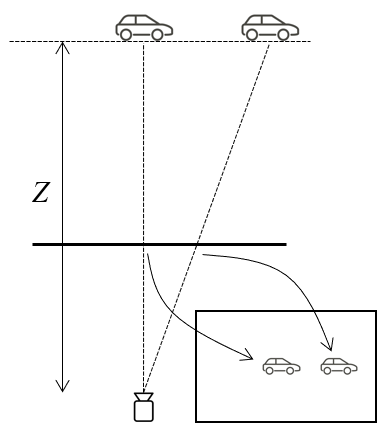}
        \caption{}
     \end{subfigure}
     \begin{subfigure}[b]{0.24\textwidth}
         \centering
         \includegraphics[width=\textwidth]{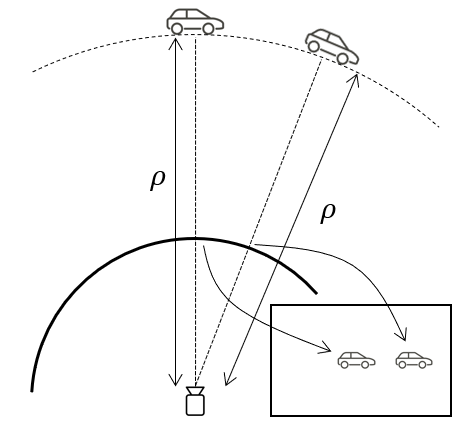}
        \caption{}
     \end{subfigure}
        \caption{In a perspective image (a), a translation with a constant $Z$-depth in the scene leads to a translated image. In contrast, with the cylindrical projection, the horizontal distance $\rho$ must be constant (that is, the object must undergo rotation around the cylinder's axis) to induce a translation in the image.}
        \label{fig:inducing_translation}
\end{figure}

A natural approach to address these issues is to rectify the images in some manner. We can immediately discard rectifying to a single planar image, as firstly, too much of the field of view would necessarily be lost thus negating the advantage of fisheye imagery and secondly interpolation and perspective artefacts would quickly dominate the rectified output. A common approach is use multi-planar rectification, whereby different portions of the fisheye image are warped to different planar images. For example, we can define a cube, and warp the image on to the surfaces of the cube. Figure \ref{fig:multilinear} shows warpings on to two such surfaces. Even here, interpolation and perspective effects are visible, and one must deal with the complexity of surface transitions. Another rectification approach is to consider warping to a cylindrical surface, as per Figure \ref{fig:cylindrical}. In such a warping, the axis of the cylinder is configured such that it is vertical to the ground. The observation is that most objects of interest in an automotive scene lie and move on an approximately horizontal plane, being the road surface. Therefore, we wish to retain the horizontal field of view, while allowing some vertical field of view to be sacrificed. This brings about an interesting combination of geometries. The vertical is per a linear perspective projection, and as such vertical lines in the scene are projected as vertical lines in the image. 
Objects that are distant or small in the image are visually like a perspective camera. It is even proposed that, through this warping, you can train a network using standard perspective cameras, and use them on fisheye imagery directly without training \cite{Plaut_2021_CVPR}. However, in the horizontal, distortion exists in the new image. Large, close objects exhibit strong distortion, sometimes even greater than in the original fisheye image. It is also interesting to consider what induces a translation in the resulting cylindrical image. As described in Figure \ref{fig:inducing_translation}, when we are dealing with a perspective camera, translation is induced when the object moves with a constant $Z$-distance from the camera. That is, on a plane that is parallel to the image plane. However, in a cylindrical image, the distance over the horizontal plane must remain constant to induce an image translation. That is, the object must undergo a rotation about the cylinder axis. In contrast, it is not clear in a raw fisheye image what object motion, if any, would induce an image translation. 


The possible warpings are essentially infinite, as we can consider any viable surface, and different warpings may prove better for certain applications and processing techniques than others. This will naturally prove to be an issue as we try to unify automotive vision techniques. With that in mind, we should start thinking instead about how to natively process fisheye imagery without warping. This is non-trivial. For convolutional techniques, for example, this necessitates revisiting how convolutions work. Work has proceeded to address this in the similar field of spherical imagery \cite{Cohen2018, coors2018spherenet, Eder2019}, but such spherical geometry is simpler to consider than native fisheye, and spherical images are, in fact, often warpings of multiple fisheye images \cite{Ho2017}.

\subsection{WoodScape dataset}

Even though this paper should be considered as part review and part positional, throughout we will offer insights into results to support our arguments. Mostly, though not ubiquitously, we will use the WoodScape Dataset \cite{yogamani2019woodscape}. It is therefore pertinent to give a description of the dataset, which was captured in two distinct geographical locations: USA and Europe. While most data were obtained from saloon vehicles there is a significant subset from a sports utility vehicle ensuring a strong mix in sensor mechanical configurations. Driving scenarios are divided across highway, urban driving, and parking use cases.
Intrinsic and extrinsic calibrations are provided for all sensors as well as timestamp files to allow synchronization of the data. Relevant vehicle's mechanical data (e.g., wheel circumference, wheelbase) are included. High-quality data are ensured via quality checks at all stages of the data collection process.
Annotation data undergo a rigorous quality assurance by highly skilled reviewers.
The sensors recorded for this dataset are listed below:
\begin{itemize} [nosep]
  \item 4x 1MPx RGB fisheye cameras ($190^\circ$ horizontal FOV)
  \item 1x LiDAR rotating at 20Hz (Velodyne HDL-64E)
  \item 1x GNSS/IMU (NovAtel Propak6 \& SPAN-IGM-A1)
  \item 1x GNSS Positioning with SPS (Garmin 18x)
  \item Odometry signals from the vehicle bus. \\
\end{itemize}
Table \ref{tab:woodscape_baseline} gives an overview of the baseline results for this dataset for three of the four Rs. Most of the metrics are standard. The relocalization metric is the percentage of instances in which the estimated pose is within a tolerance of 2{\degree} in orientation and 0.05m in position.


\begin{table}[t]
\centering
\caption{Summary of baseline results of 4Rs on our WoodScape dataset. } 

\begin{adjustbox}{width=0.5\textwidth}
\begin{tabular}{lccc}
\hline
\textbf{Task} & \textbf{Model} & \textbf{Metric} & \textbf{Value} \\ \hline
\multicolumn{4}{c}{\textbf{Recognition }}\\
\hline
Segmentation & ENet \cite{paszke2016enet} & IoU & 51.4 \\
2D Bounding Box & Faster R-CNN \cite{ren2015faster} & mAP (IoU\textgreater{}0.5) & 31 \\
\multirow{1}{*}{Soiling Detection} & \multirow{1}{*}{ResNet10 \cite{he2016deep}} & Category (\%) & 84.5 \\ \hline
\multicolumn{4}{c}{\textbf{Reconstruction }}\\ \hline
Depth Estimation & Eigen \cite{eigen2015predicting} & RMSE & 7.7 \\
Motion Segmentation & MODNet \cite{siam2018modnet} & IoU & 45 \\
\multirow{2}{*}{Visual Odometry} & \multirow{2}{*}{ResNet50 \cite{he2016deep}} & Translation (\textless{}5mm) & 51 \\ 
                                 &                                             & Rotation (\textless{}0.1\degree) & 71 \\
\hline
\multicolumn{4}{c}{\textbf{Relocalization }} \\ \hline
Visual SLAM & LSD SLAM \cite{engel2014lsd} & Relocalization (\%) &  61 \\
\hline
\end{tabular}
\end{adjustbox}
\vspace{1ex}
\label{tab:woodscape_baseline}
\end{table}

\section{System Architecture Considerations} \label{sec:arch}

A significant consideration in the design of automotive computer vision, in particular the pipelining, is the constraints of embedded systems in which multiple cameras and multiple computer vision algorithms must run in parallel. It is therefore useful to give a brief overview to understand the constraints better, though readers are referred to \cite{heimberger2017computer} for a more detailed review of these considerations. Perhaps the most important component to consider is the System-on-Chip (SoC), and typically the first step in designing a commercial automotive camera system is the selection of the SoC for embedded systems, based on criteria including performance (Tera Operations Per Second (TOPS), utilization, bandwidth), cost, power consumption, heat dissipation, high to low end scalability and programmability. The SoC choice provides the computational bounds in the design of algorithms. As computer vision algorithms are compute intensive, Automotive SoCs have a lot of dedicated hardware accelerators for image signal processing, lens distortion correction, dense optical flow, stereo disparity, etc. In computer vision, deep learning is playing a dominant role in various recognition tasks and gradually for geometric tasks, like depth \cite{Kumar2018CVPRWorkshop} and motion estimation \cite{wang2017deepvo}. The progress in CNN has also led to the hardware manufacturers typically including a custom hardware intellectual property core to provide a high throughput of over $10$ TOPS \cite{mittal2019survey}.

\subsection{Pipeline Stages}

To maximize the performance of processing hardware, it is advantageous to consider embedded vision in terms of processing stages, and to consider shared processing at each processing stage. In this manner, expensive early operations are shared amongst later processing stages that are closer to application layers. In Figure \ref{fig:4R_arch}, we show an example of a 4R architecture split into pipeline stages.

\definecolor{heavygreen}{HTML}{00BF00}
\begin{figure*}[tb]
    \centering
    \includegraphics[width=0.9\linewidth]{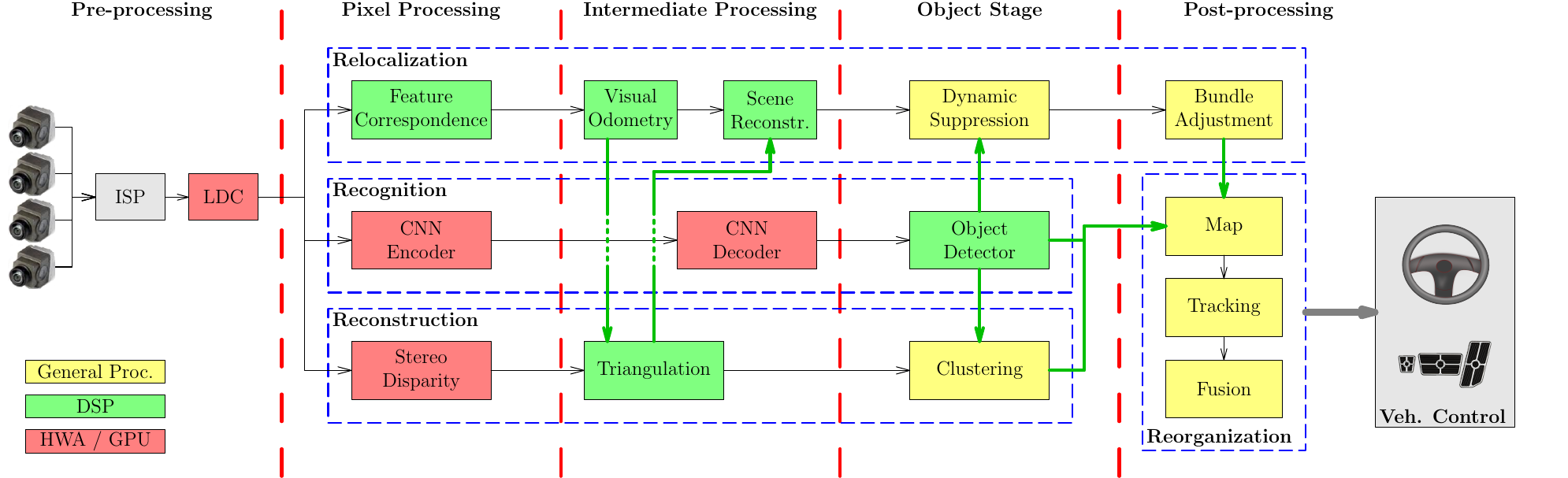}
    \caption{Example of a high level architecture proposal considering a 4R framework. The 4Rs are shown with each of the processing stages. Useful communication between each of the 4Rs is shown as \textcolor{heavygreen}{green} arrows. Each block has an example target processing unit, being Hardware Accelerator (HWA) or GPU, Digital Signal Processor (DSP), or general-purpose processing unit, such as ARM. LDC = Lens Distortion Correction, ISP = Image Signal Processor.}
    \label{fig:4R_arch}
\end{figure*}

\subsubsection{Pre-processing}

The pre-processing stage of the pipeline can be thought of as the processing that prepares the data for computer vision. This consists of the Image Signal Processing (ISP) steps, such as White Balance, Denoise, Color Correction and Color Space Conversion. For a full discussion on ISP and the tuning of ISP for computer vision tasks in an automotive setting, the reader is referred to \cite{yahiaioui2019isp}. ISP is usually done by hardware engines, e.g., as part of the primary SoC. It is rarely done in software, as there is a vast amount of pixel level processing to be completed. Methods are being proposed to automatically tune the hyperparameters of ISP pipelines to optimize the performance of computer vision algorithm \cite{yahiaioui2019isp, mosleh2020isp}. It should be noted that methods are being proposed to simplify the ISP pipeline for visual perception \cite{Buckler_2017_ICCV}.

\subsubsection{Pixel Processing Stage}

Pixel processing can be considered as those parts of a computer vision architecture that \textit{touch the image} directly. In classical computer vision, these would be algorithms such as edge detection, feature detection, descriptors, morphological operations, image registration, stereo disparity and so on. Readers are referred to any one of a number of textbooks for further information on these steps, such as \cite{Szeliski2010}. In neural networks, this would equate with the early layers of a CNN encoder. The processing at this stage is dominated by relatively simple algorithms that must run on potentially millions of pixels many times each second. That is, the computational cost is associated with the fact that these algorithms may run many millions of times each second, rather than the complexity of the algorithms themselves. Processing hardware at this stage is typically dominated by Hardware Accelerators and GPUs, though some elements may be suitable for DSP.

\subsubsection{Intermediate Processing Stage}

As the name suggests, the intermediate processing stage bridges the gap from the pixels to the object detection stage. Here, the amount of data to process is still high, but significantly lower than the pixel processing stage. This may include steps such as estimating vehicle motion through visual odometry, stereo triangulation of a disparity map, and the general feature-wise reconstruction of a scene. We would also include CNN decoders at this stage of the pipeline. Processing hardware at this stage would typically be digital signal processors.

\subsubsection{Object Processing Stage}

The object processing stage is where higher level reasoning is incorporated. It is here that we may cluster point clouds to create objects, where objects are classified, and where, through said reasoning, we can apply algorithms to suppress relocalization on movable objects. The processing at this stage is dominated by more complex algorithms but operating on fewer data points. In terms of hardware, it is often suitable to run these on general purpose processing units, such as ARM, though digital signal processors would commonly be utilized as well.

\subsubsection{Post-processing} 

Finally, we have the post processing stage, which could also be termed the \textit{global stage} of processing. It is here that we persist data temporally and spatially. As we can have long temporal persistence and large spatial maps, the overall goal of the preceding stages is to minimize the amount of data reaching this stage while maintaining all the pertinent information that will finally be used for vehicle control. In this stage, we would include steps such as bundle adjustment, map building, high level object tracking and prediction and fusion of the various computer vision inputs. As we are dealing with the highest level of reasoning in the system, and ideally with the lowest amount of data points, general purpose processing units are typically desirable here.

\section{4R Components} \label{sec:components}

It is useful to begin the detailed description of the 4Rs with an example, and so, in Figure \ref{fig:4R_arch}, we show at a high level what a computer vision architecture for autonomous driving might look like. Four video streams from an automotive surround-view system (Figure \ref{fig:svs}) are passed to an SoC. Each algorithmic block is mapped to one of a set of available processing units that are typically available on high-end automotive vision SoCs, being hardware accelerator or graphical processing unit, digital signal processor, and general-purpose processing core (e.g., ARM). Each of the 4R pipelines shows one of the standard algorithms, being visual SLAM for relocalization, CNN for object detection and motion stereo for reconstruction. Reorganization shows a typical map, tracking and fusion pipeline.

What is of interest is the possible links between the 4Rs, even in such a standard system. For example, Visual Odometry is a natural part of any SLAM pipeline but can be reused in a motion stereo \cite{szeliski1999motionstereo} context for dense triangulation, and equally, the stereo triangulation can be used for the scene reconstruction component of visual SLAM, which is really a seed for bundle adjustment. Robust SLAM can only be achieved in scenes that are dominated by dynamic objects if motion segmentation is readily available \cite{saputra2018dynslam}. CNNs provide options on moving object detection \cite{yahiaoui2019fmodnet}, potentially as part of a multi-task network \cite{sistu2019neurall}, and thus can be used to suppress issues in the relocalization pipeline associated with dynamic objects. Motion segmentation can also be used to suppress issues with incorrect triangulation in reconstruction, and as such can be included in the clustering stage of the reconstruction pipeline to suppress false detections. Finally, naturally, the output of the object detector can be directly input into the sensor map.

Now that we have briefly discussed a simple example of how the 4Rs architecture might work, we will describe in more detail what the 4Rs are.

\subsection{Recognition} 

The \textit{Recognition} task identifies the semantics of the scene via pattern recognition. 
In automotive, the first successful application was pedestrian detection which was performed using a combination of hand-designed features like Histogram of Oriented Gradients and a machine learning classifier like Support Vector Machines (for example \cite{Bauer2010} and many others). 
Recently, this task has been dominated by CNNs, which have demonstrated remarkable performance leaps for various computer vision tasks in object recognition applications \cite{hassaballah2020deep}. However, this comes at a cost. Firstly, automotive scenes are very diverse, and the system is expected to work across countries as well as varying weather and lighting conditions, and thus one of the main challenges is to build an effective dataset which covers diverse aspects \cite{dataset_visapp19}. Secondly, CNNs are computationally intensive, typically requiring dedicated hardware accelerators or GPUs (in contrast to classical machine learning approaches that are feasible on general purpose computation cores). As such, efficient design techniques are critical to be incorporated in any design \cite{briot2018analysis,das2019design}. Finally, while CNNs are well studied for rectilinear images, as mentioned previously, the assumption of translation invariance is broken in fisheye images, which pose additional challenges as discussed in \cite{kumar2020unrectdepthnet}. In particular, standard bounding box object detection representation breaks for fisheye images \cite{rashed2021generalized}, though special consideration should also be given to the design of semantic segmentation approaches for fisheye \cite{Deng2017}.

In our example recognition pipeline, a multi-task deep learning network for identifying objects based on their appearance patterns is proposed. It comprises of three tasks, namely bounding box objection detection (pedestrians, vehicles, and cyclists), semantic segmentation (road, curbs, and road markings) and lens soiling detection (opaque, semi-transparent, transparent, clear). Object detection and semantic segmentation are standard tasks and for more implementation details the reader is referred to our FisheyeMultiNet paper \cite{maddu2019fisheyemultinet}. One of the challenges is to balance the three tasks’ weights during training phase as one task may converge faster than the others \cite{leang2020dynamic}. Additional auxiliary tasks like end-to-end driving which do not have associated annotation costs can aid the training of expensive annotation tasks such as segmentation \cite{chennupati2019auxnet}.

\begin{figure}[ht]
    \centering
    \includegraphics[width=\linewidth]{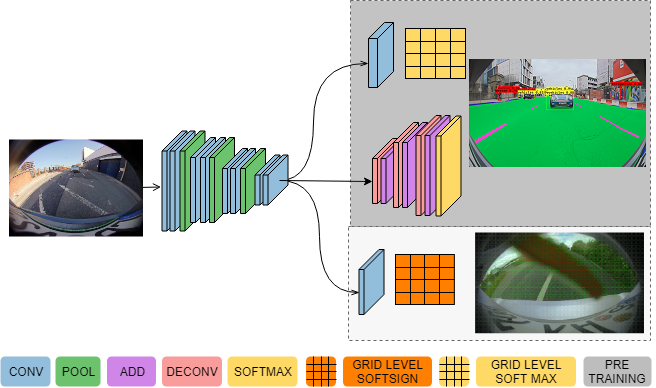}
    \caption{Illustration of multi-task Recognition architecture comprising of object detection, semantic segmentation and soiling detection tasks.}
    \label{fig:recog}
\end{figure}

Fisheye cameras are mounted relatively low on a vehicle ($\sim$0.5 to 1.2m above ground) and are susceptible to lens soiling due to road spray from other vehicles or water from the road. Thus, it is vital to detect soiling on the camera lens to alert the driver to clean the camera or to trigger a cleaning system. The soiling detection task and its usage for cleaning and algorithm degradation is discussed in detail in SoilingNet \cite{uvrivcavr2019soilingnet}. A closely related task is desoiling where the soiled areas are restored through inpainting \cite{uricar2019desoiling}, but these desoiling techniques remain in the domain of visualization improvements rather than usage for perception for now. It is an ill-defined problem as it is not possible to predict behind the occlusion, though this can be improved by leveraging temporal information.
As the CNN processing capacity is limited on the low power automotive ECU, we make use of multi-task architecture where majority of the computation is shared in the encoder as illustrated in Figure \ref{fig:recog}.

\subsection{Reconstruction} 

As mentioned already, {\em Reconstruction} means inferring scene geometry from a video sequence. This typically means, for example, estimating a point cloud or voxelized representation of a scene. However, we can also consider a temporal aspect of this -- if the object is moving, we wish to know its vector of motion.

The first aspect, the reconstruction of static objects, is traditionally done using approaches such as motion stereo \cite{szeliski1999motionstereo} or triangulation in multi-view geometry \cite{kukelova2019triang}. In the context of designing a depth estimation algorithm, a brief overview of how humans infer depth is given in \cite{saxena2007monodepth} with useful further referencing. There are four basic approaches to inferring depth: monocular visual cues, motion-parallax, stereopsis, and depth from focus. Each has its equivalent in computer vision.

Based on earlier theoretical work by Marr \& Poggio \cite{marr1979stereo}, Grimson provided a computational implementation of stereo vision in the early 1980s \cite{grimson1981stereo}. Since then, work has continued on stereo vision (see \cite{davies2005stereo} for an overview of early works). However, stereo systems do not achieve ubiquitous deployment on vehicles, and as such, monocular motion-parallax methods remain popular in automotive research. Computationally, depth from motion parallax is traditionally done through feature triangulation \cite{kukelova2015radial}, but motion stereo has also proven popular \cite{hirata2019depth}. Neural network approaches to extracting pixel-wise depth from moving cameras have been successful recently \cite{kumar2018monocular}.

Perhaps one of the more interesting approaches is the use of monocular cues for depth. These are things like texture scale change, occlusions (if A occludes B, then B must be behind A), shading and lighting, object scale (small vs far away), etc. As noted in \cite{saxena2007monodepth}, such monocular cues require contextual interpretation. To work well, they require knowledge of the entire image, and patch-based approaches generally fail. Van Dijk and Croon \cite{dijk2019monodepth} have shown that, for four publicly available mono-depth networks at least, neural networks learn a correlation between vertical position of the object in the image and the depth to the object. Single-view (one camera, one frame) approaches are extremely desirable, as they remove the need for the camera to move for depth extraction. However, they have proven not to be robust enough \cite{dijk2019monodepth}, though they may be useful still in certain scenarios (when the car is still may be better than no estimate).

Considering fisheye imagery adds significant complexity to the reconstruction task. Most work in multi-view geometry, stereo vision and depth estimation in general assumes a planar perspective image of the scene, i.e., that projective geometry provides a good model of the image. Traditional stereo approaches add a further restriction that the epipolar lines in the image must be horizontal. However, this is rarely the case with real cameras where lens distortions exist thus breaking the planar projection model. It is generally addressed through calibration and rectification of the image. For fisheye imagery where the lens distortion is extreme, though, it is not feasible to maintain the wide field of view in rectification.

Several approaches have been proposed to address fisheye stereo depth estimation. A common approach is multi-planar rectification, in which the fisheye image is mapped to several perspective planes \cite{Gao2017Fisheye}. However, any planar rectification, even with multiple planes, suffers from significant resampling distortion, as discussed earlier. To minimize this resampling distortion, rectification to non-planar images has been proposed. Some approaches warp to different image geometries that maintain the stereo requirement of epipolar lines being straight and horizontal \cite{Esparza2014}. Still other approaches bypass the requirement for epipolar lines to be horizontal. For example, the plane sweep method \cite{Collins1996, Gallup2007} has more recently been applied to fisheye \cite{Hane2014}. A related issue with any resampling of the fisheye image is that the noise function is distorted by the resampling process, which is a problem for any method that attempts to minimize a reprojection error (e.g., the widely used optimal triangulation method \cite{hartley1997triangulation}). Kukelova et al. \cite{kukelova2019triang} address this using an iterative technique for standard field of view cameras that minimizes reprojection error while avoiding undistortion. However, this approach depends on a specific camera model, and as such is not directly applicable to fisheye cameras.

The second aspect of reconstruction is the extraction of moving objects from the video sequence (motion segmentation). 3D-reconstruction of dynamic objects results in position inaccuracy in the global sense, as triangulation assumptions are broken. Typical attempts to reconstruct the geometry of an object under motion requires image motion segmentation, relative fundamental matrix estimation and reconstruction (with scale/projective ambiguity). Of course, significant advances have been made. For example, using Multi-X \cite{barath2018multix} the first two steps can essentially be combined, as the segmentation can be done based on the fundamental matrix estimation. However, such approaches tend to be either computationally too expensive or not robust enough for embedded automotive applications. Additionally, scale must be resolved for such reconstruction, and deformable objects (such as pedestrians) can have different fundamental matrices for different parts of the body. Therefore, in automotive, the task of dynamic object detection is usually simply motion segmentation.

Klappstein et al. \cite{klappstein2007} describe a geometric approach to motion segmentation in the automotive context. This work is significantly extended to the surround-view camera case by Mariotti and Hughes \cite{mariotti2020motion}. However, in both cases the geometry cannot perfectly distinguish all types of moving feature. That is, there is a class of object motion that makes associated features indistinguishable from static features. Thus, a global or semi-global approach must be taken. In traditional approaches, this is done by grouping optical flow vectors with similar properties to ones that are classed as under motion. CNNs offer globality in a more native way \cite{yahiaoui2019fmodnet}, and even offer the potential for instance motion segmentation, though this has yet to be extended to the fisheye case \cite{Mohamed2021}. However, as with the static object reconstruction, the results from \cite{yahiaoui2019fmodnet} seem to indicate that it is performing recognition rather than geometric motion estimation, as still pedestrians are often classed as under motion. It is therefore likely that a much-improved overall motion segmentation will be obtained by incorporating the geometric constraints of \cite{mariotti2020motion} with the more global CNN approach of \cite{yahiaoui2019fmodnet}, though this is certainly non-trivial and remains work in progress.

Generally, a key input into motion segmentation is knowledge of the motion of the camera. That is, the essential matrix of the camera (or fundamental matrix in the uncalibrated case) must be known. This is assumed in \cite{klappstein2007} and \cite{mariotti2020motion}. This can be achieved in a couple of ways. Firstly, we can directly use signals on the vehicle network, such as steering angle and wheel velocities, to estimate the motion of the vehicle (e.g., as discussed in \cite{Brunker2019}), and thus the motion of the cameras. Alternatively, visual approaches to estimate the motion directly from image sequences can be employed \cite{aqel2016}. An alternative to the explicit estimation of the motion of the camera is to model the background motion in the image. It has been proposed to use an affine model of background motion \cite{Bugeau2009motion}. However, this assumes that the background is distant or approximately planar, and that radial distortion is absent or negligible. The latter is clearly not the case with automotive fisheye, but perhaps not as obviously, nor is the former. It is obvious that the background for a typical automotive scene (being the road surface, buildings, etc.) cannot be modelled as a single plane, nor is it very distant.

\begin{figure}[ht]
    \centering
    \includegraphics[width=\linewidth]{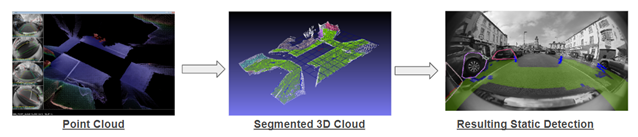}
    \includegraphics[width=\linewidth]{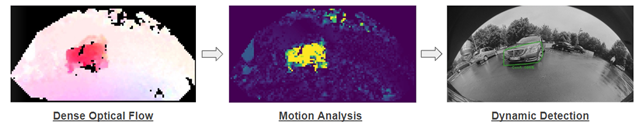}
    \caption{An example of a reconstruction pipeline with some sample outputs at different processing stages. Top row shows the static pipeline and bottom row shows the dynamic pipeline.}
    \label{fig:recon}
\end{figure}

Figure \ref{fig:recon} shows an example of different reconstruction stages, including dense motion stereo, 3D point cloud and a clustering of static obstacle, alongside a dense optical flow based motion segmentation. While the use of fisheye images certainly has an impact on design decisions and may be considered a problem that is not yet fully solved from a theoretical standpoint, it is clear that in practice many of the techniques for fisheye discussed above give acceptable results, depending on specific applications.

\subsection{Relocalization} 

Visual Simultaneous Localization and Mapping (VSLAM) is a well-studied problem in robotics and autonomous driving. There are primarily three types of approaches namely (1) Feature based methods, (2) Direct SLAM methods and (3) CNN approaches. Feature based methods make use of descriptive image features for tracking and depth estimation \cite{kumar2019fisheyedistancenet} which results in sparse maps. MonoSLAM \cite{Davison2007}, Parallel Tracking and Mapping (PTAM) \cite{klein07parallel} and ORBSLAM \cite{mur2015orb} are seminal algorithms of this type. Direct SLAM methods work on the entire image instead of sparse features to aid building a dense map. Dense Tracking and Mapping (DTAM) \cite{Newcombe2011} and Large-Scale Semi Dense SLAM (LSD-SLAM) \cite{caruso2015large} are the popular direct methods which are based on minimization of photometric error. CNN based approaches are relatively less mature for Visual SLAM problems and they are discussed in detail in \cite{milz2018visual}.

Mapping is one of the key pillars of autonomous driving. Many first successful demonstrations of autonomous driving (e.g., by Google) were primarily reliant on localization to pre-mapped areas. HD maps such as TomTom RoadDNA \cite{TomTomRoadDNA} provide a highly dense semantic 3D point cloud map and localization service for majority of European cities with a typical localization accuracy of 10 cm. When there is an accurate localization, HD maps can be treated as a dominant cue, as a strong prior semantic segmentation is already available, and it can be refined by an online segmentation algorithm \cite{ravi2018real}. However, this service is expensive as it requires regular maintenance and upgrades of various regions in the world. Due to privacy laws and accessibility, such a commercial service cannot be used in many situations and a mapping mechanism must be built within a vehicle’s embedded system. For example, a private residential area cannot be mapped legally in many countries, such as Germany \cite{luo2019localization}.

Visual SLAM (VSLAM), in the automotive context, consists of building a map of the environment surrounding the vehicle while simultaneously estimating the current pose of the car within that map \cite{Singandhupe2019slam}. One of the key tasks of VSLAM is the localization of the vehicle against a previously recorded trajectory \cite{Kasyanov2017slam}. A trained trajectory is typically represented by a group of key poses surrounded by landmarks spanned from the vehicle's origin to destination positions. These landmarks are represented using robust image features that are unique in the captured images.

\begin{figure*}[ht]
    \centering
    \includegraphics[width=0.7\linewidth]{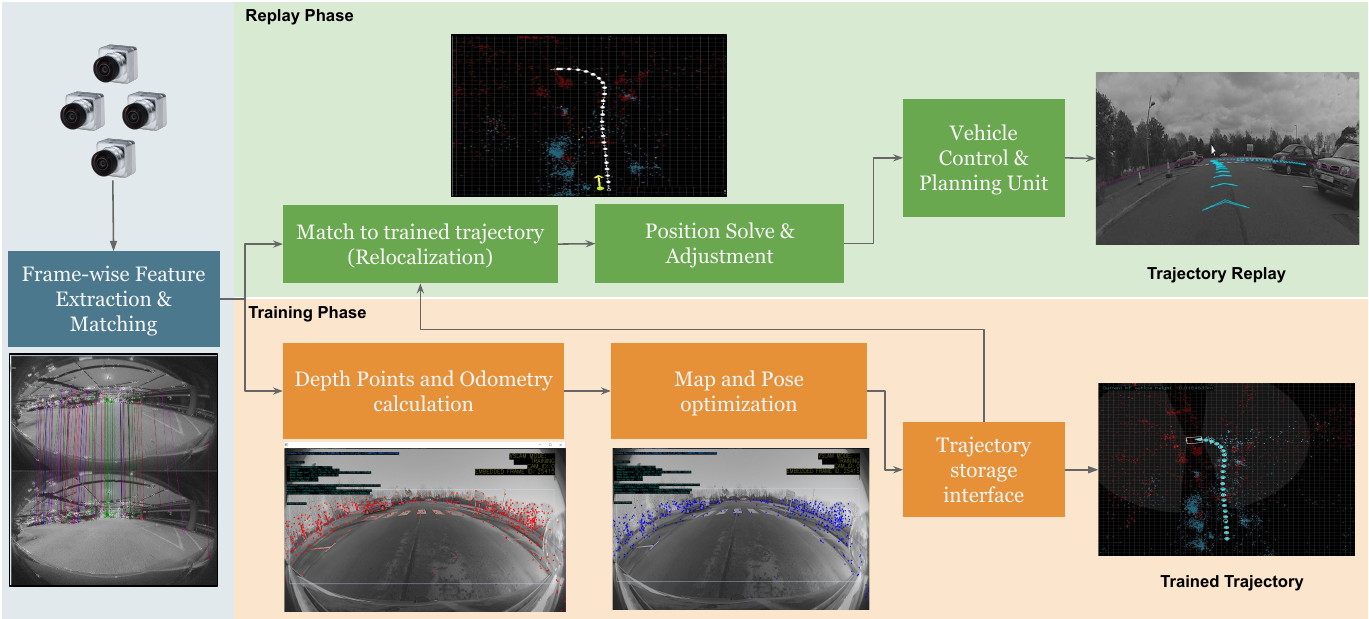}
    \caption{Relocalization pipeline with intermediate outputs. Reproduced from our paper \cite{tripathi2020trained}. }
    \label{fig:reloc_pipeline}
    \vspace{-0.5cm}
\end{figure*}

A classical feature-based relocalization pipeline is shown in Figure \ref{fig:reloc_pipeline}. In feature-based SLAM, the first step is the extraction of salient features. A salient feature in an image could be a region of pixels where the intensity changes in a particular way, such as an edge, a corner or a blob \cite{lowe1999features, rublee2011features, Alcantarilla2011features}. To estimate landmarks in the world, tracking is performed, wherein two or more views of the same features can be matched. Once the vehicle has moved enough, VSLAM takes another image and extracts features. The corresponding features are reconstructed to get their coordinates and poses in real world. These detected, described, and localized landmarks are then stored in persistent memory to describe the relative position of the vehicle for a trajectory. If the vehicle returns to same general location the live feature detections are matched against the stored landmarks to recover the vehicle's pose relative to the stored trajectory.

\subsection{Reorganization} 

\begin{figure}[ht]
  \centering
    \includegraphics[width=0.8\linewidth]{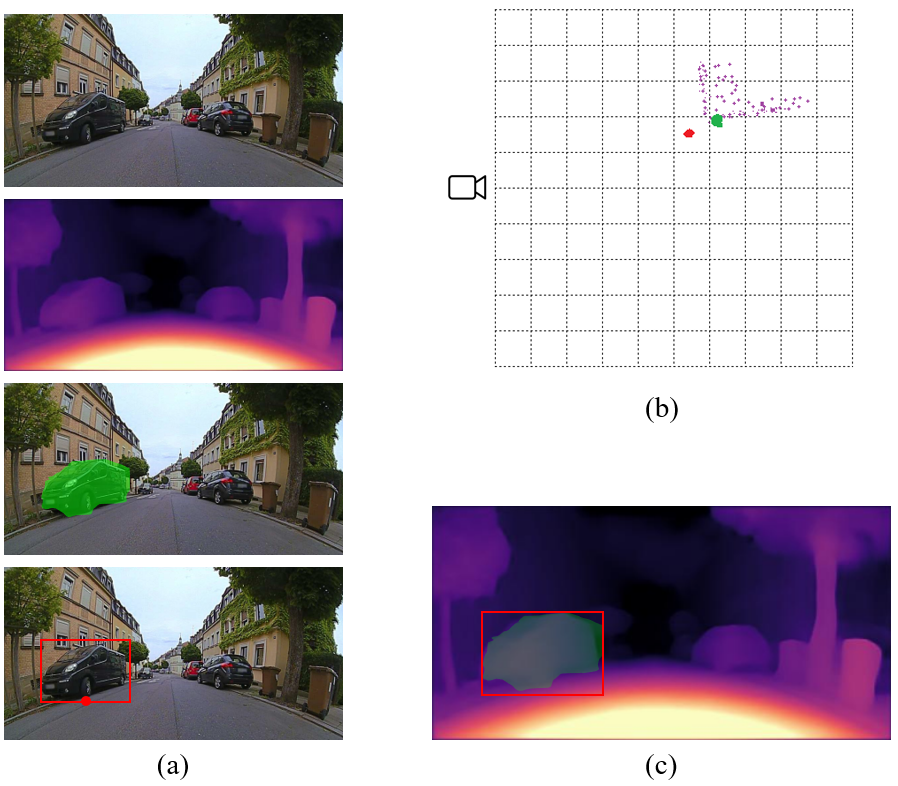}
    \caption{Two fusion paradigms. (a) shows the inputs, being a monocular depth estimation, motion segmentation and bounding box. (b) shows the world coordinate conversion of these detections, generating a point cloud for the monocular depth and the projection of a reference point for each of the motion segmentation and bounding box detection. It is not clear how we could associate all the detections in a fusion system. In contrast, (c) shows all the detections in a single image plane, where it is intuitive that such fusion is almost trivial (a simple metric such as overlap would suffice).}
    \label{fig:fusion}
    \vspace{-0.5cm}
\end{figure}

\begin{figure}[ht]
  \centering
    \includegraphics[width=\linewidth]{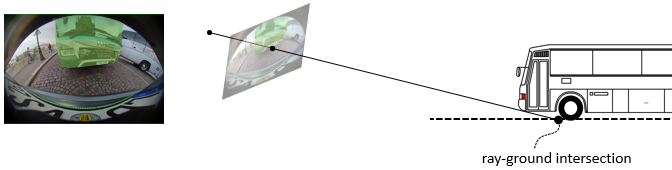}
    \caption{Even if we had a hypothetical perfect semantic segmentation, if the object does not intersect the ground at the point of interest (e.g., the front of the bus), then the ray projection-based estimation of the world coordinate system object position will necessarily be in error.}
    \label{fig:ray_ground}
\end{figure}

Reorganization performs three functions - 1) Fusion of Recognition and Reconstruction, 2) Mapping of objects in a centralized world co-ordinate system across cameras and 3) Temporal tracking of objects. Although it would be possible for the recognition and reorganization blocks to feed directly into the environment map, we contend there are distinct advantages to implementing some fusion at the vision layer.
Let us consider this first with an example. As shown in Figure \ref{fig:fusion}, let us assume we have a system that has a monocular depth estimation, motion segmentation and bounding box vehicle detection. A classical approach to fusing this information is to convert all the data into a world coordinate system, and then associate and fuse the data. This type of approach has advantages. Some automotive sensors, such as laser scanner, provide native Euclidean data, and a fusion system based on such a Euclidean map makes inclusion of these additional sensors easy. However, camera-based detection accuracy will always suffer with the conversion to a Euclidean map.
Projections from the image domain to the world domain are known to be error prone, as they are subject to errors from poor calibration, flat ground assumptions, variations in footpoint detection, pixel density and imperfect camera models. Even consider the case in which we have a perfect semantic segmentation, as in Figure \ref{fig:ray_ground}. If the object does not actually touch the ground at the point of interest, then there will be significant error with the flat-ground assumption for projection to a world coordinate system.


\begin{figure*}[ht]
    \centering
    \includegraphics[width=0.7\linewidth]{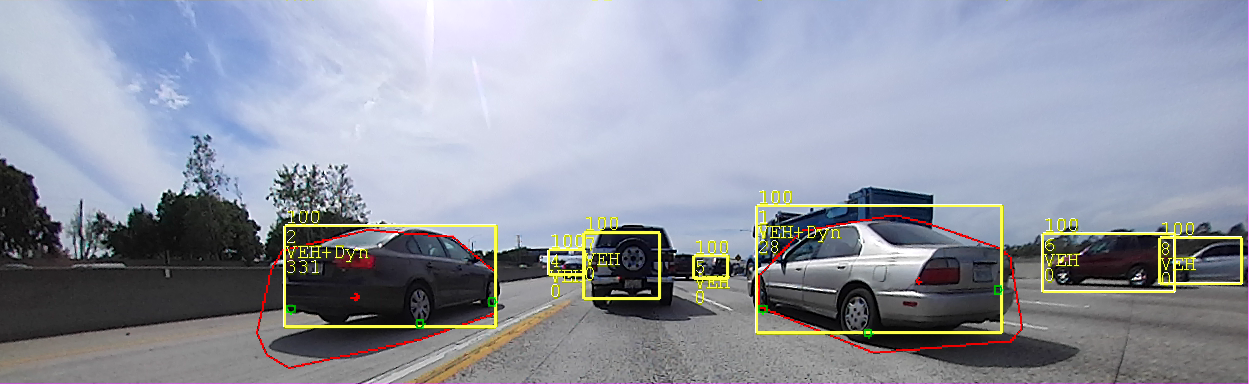}
    \caption{Reorganization combines Reconstruction's dynamic object detection output (red polygon) and Recognition's object detection (yellow box).}
    \label{fig:reorg_mod}
\end{figure*}

However, detections in the image domain, prior to projections to the world, are not subject to such error and therefore association of detections from different vision algorithms in the image domain is more robust.
In fact, simple detection overlap measures typically prove robust. Figure \ref{fig:reorg_mod} shows an implementation of the image-based fusion of a CNN-based vehicle detection and an optical flow-based motion segmentation. Even though significant error exists in the motion segmentation, the fusion successfully classifies the detected objects as both vehicle and dynamic.

\begin{figure}[ht]
  \centering
    \includegraphics[width=\linewidth]{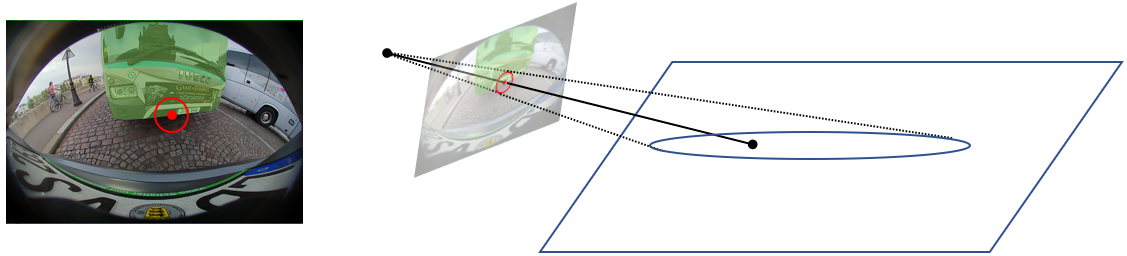}
    \caption{Any projection from a fisheye image to the world coordinate system will necessarily induce a non-linear distortion of the measurement noise model, due to the non-linearities of the fisheye unprojection and flat ground intersections. Measurement noise is no longer zero-mean, Gaussian, nor even symmetric.}
    \label{fig:noise_projection}
\end{figure}

One must also consider how the distortion correction impacts measurement noise. Many common algorithms for fusion and tracking, such as the Kalman filter or the particle filter, begin with the assumption of mean-zero, Gaussian noise. For interest point measurement in computer vision (e.g., an image feature or a bounding box footpoint estimate), this is commonly considered to be a valid assumption. However, the fisheye undistortion and ground plane projection process distorts this noise model (Figure \ref{fig:noise_projection}). Addressing this is additionally complicated by the fact that the distortion of the measurement noise is dependent on the location of the interest point in the image and the position of the camera relative to the road surface.

So far in this section, we have predominantly focused on fusion. However, tracking the objects in the scene will suffer from similar artefacts as fusion if we attempt to do this in the world coordinate system. Following detection of an object, one must associate that detection with a previous detection of the same object. If we rely on something like the prediction of the footpoint of an object in world coordinates (per Figure \ref{fig:noise_projection}), noise will generally mean that the association is very difficult. However, in image coordinates, geometric measures of overlap (e.g., intersection-over-union) are more robust against noise, and the association problem, as with the fusion problem, is significantly simpler. Naturally, we don't intend to give a complete survey of multi-object tracking. The reader is instead referred to \cite{CIAPARRONE202061}, which discusses classical as well as more modern neural network approaches.

The two common mapping paradigms in automotive perception are vector maps \cite{Baizid2016} and occupancy grids \cite{thrun2005book}, and sensor fusion generally occurs in such a map (that is, fusion between different sensor types, such as camera, laser scanner, radar, etc.). A survey of sensor fusion is beyond the scope of this paper; we would refer the reader to \cite{Wang2019} for a very comprehensive review of automotive sensor fusion modalities. Ultimately, we will want the vehicle to make control decisions based on image perception. Given that the vehicle exists in a Euclidean space, and planning and control decisions are generally made on Euclidean data represented by the perception map, map information must be generated from image perception at some point. By fusing detections in the image space, we can, for example, smooth the detections of object footpoints. This will mean a better localization in world coordinates. More importantly, however, we can more easily fuse object detection with depth information from an image. Using Figure \ref{fig:fusion} as an example, if we have an accurate depth estimation in our image space, we can attach the semantic labels to the depth prior to generating the map information.

\subsection{Discussion}
Overall, we argue that the 4R approach provides a localized semantic-geometric representation of the vehicles environment. By \textit{localized semantic-geometric} representation of the vehicles environment, we mean \textit{localized}: that the 4R processing pipeline provides information about where the vehicle is (can be globally or against pre-learned trajectories), \textit{geometric}: information about the spatial relationship between the vehicle and obstacles in its local environment, and \textit{semantic}: the obstacles will be recognized as belonging to a class of obstacle.

\section{System Synergies} \label{sec:synergies}

In this section, we will discuss system synergies. We will look at how Relocalization, Reconstruction and Recognition tasks can support one another, and we will describe the importance of dual sources of detection in providing redundancy in safety critical applications. 

\subsection{Recognition and Reconstruction}

\begin{figure}[ht]
  \captionsetup{singlelinecheck=false, font=small,  belowskip=-10pt}
  \centering
    \includegraphics[width=0.8\linewidth]{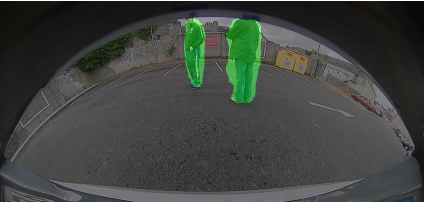}
    \caption{Learning geometry can have a heavy reliance on recognition - static pedestrians detected as moving \cite{yahiaoui2019fmodnet}.
    }
    \label{fig:fisheyemodnet_fps}
\end{figure}

As already mentioned, depth estimation is important in geometric perception application. In addition to previous material already discussed, the current state of-the-art are neural network-based methods \cite{Fu-2018, Zhang-2019}, learnable in a self-supervised manner through reprojection loss \cite{Zhou-2017}. 
It has been shown that state-of-the-art single frame attempts at monocular depth estimation typically results in recognition tasks \cite{tatarchenko2019monodepth}, and then using cues such as vertical position in the image to infer depth \cite{dijk2019monodepth}. Moving object detection appears to have a heavy reliance on recognition as well. This is evidenced by the fact that both \cite{siam2018modnet} and \cite{yahiaoui2019fmodnet} show false positives on static objects that are commonly moving (pedestrians, for example - see Figure \ref{fig:fisheyemodnet_fps}). This does not, in any way, reduce the importance of such attempts. Rather, it points to a very deep connection between recognition and reconstruction, and that from one, you can infer the other.

\begin{figure}[ht]
  \captionsetup{singlelinecheck=false, font=small,  belowskip=-10pt}
  \centering
    \includegraphics[width=0.8\linewidth]{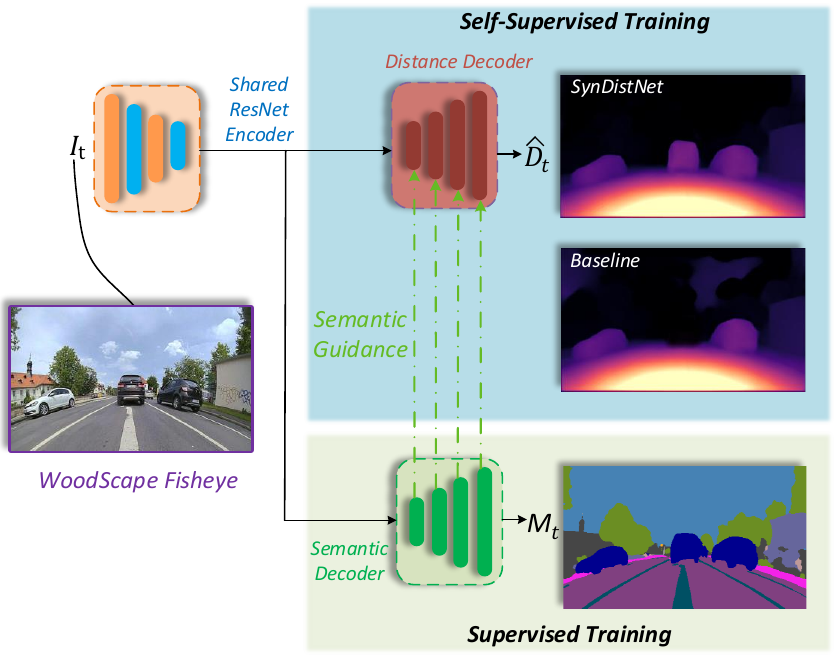}
    \caption{Overview over the joint prediction of distance and semantic segmentation from a single input image \cite{kumar2021segdepth}. 
    }
    \label{fig:recog_recon}
\end{figure}

When bounding box pedestrian detection was state of the art, before semantic and instance segmentation, most researchers in automotive pedestrian detection will have considered encoding a depth based on the height of the bounding box, or the vertical position of the pedestrian in the image. This is discussed in detail in \cite{dijk2019monodepth}. However, it is somewhat intuitive that recognition based on deep neural networks can lead to object depth, especially as the accuracy of neural networks improves. Recent work demonstrates the validity of joint learning of semantic labels and depth \cite{Lin2019}. For example, in \cite{kumar2021segdepth}, it is shown that, for monocular depth estimation, adding semantic guidance in each of the distance decoder layers (per Figure \ref{fig:recog_recon}) improves performance at edges of objects, and even returns reasonable distance estimates for dynamic objects. Table \ref{tab:recog_recon} shows an extract of the results from our work in \cite{kumar2021segdepth}, in comparison to other mono-depth approaches.

\begin{table*}[t]
\captionsetup{belowskip=-8pt, font= small, singlelinecheck=false}
\caption{Extract of our results presented in \cite{kumar2021segdepth}. Comparison of the SynDistNet network, which combines semantic segmentation and depth estimation in a single network, with other monocular methods. KITTI dataset \cite{geiger2013vision} is used in all cases. 
}
\renewcommand{\arraystretch}{0.87}
\centering
{
\small
\setlength{\tabcolsep}{0.3em}
\begin{tabular}{l|cccccccc}
\toprule
\textbf{Method} & Resolution  & Abs Rel & Sq Rel & RMSE & RMSE$_{log}$ & $\delta<1.25$ & $\delta<1.25^2$ & $\delta<1.25^3$ \\
\cmidrule(lr){3-6} \cmidrule(lr){7-9}
\multicolumn{2}{c}{} & \multicolumn{4}{c}{lower is better} & \multicolumn{3}{c}{higher is better}\\
\toprule
EPC++~\cite{Luo2019a} & 640 x 192 & 0.141 & 1.029 & 5.350 & 0.216 & 0.816 & 0.941 & 0.976 \\
Monodepth2~\cite{Godard2019} & 640 x 192 & 0.115 & 0.903 & 4.863 & 0.193 & 0.877 & 0.959 & 0.981 \\
PackNet-SfM~\cite{Guizilini2020a}  & 640 x 192 & 0.111 & \textbf{0.829} & 4.788 & 0.199 & 0.864 & 0.954 & 0.980 \\
FisheyeDistanceNet~\cite{kumar2019fisheyedistancenet} & 640 x 192 & 0.117 & 0.867 & 4.739 & 0.190 & 0.869 & 0.960 & 0.982 \\
\textbf{SynDistNet}~\cite{kumar2021segdepth} & 640 x 192 & \textbf{0.109} & 0.843 & \textbf{4.594} & \textbf{0.186} & \textbf{0.878} & \textbf{0.968} & \textbf{0.986} \\
\cmidrule{1-9} 
Monodepth2~\cite{Godard2019}                 & 1024 x 320 & 0.115 & 0.882 & 4.701 & 0.190 & 0.879 & 0.961 & 0.982 \\
FisheyeDistanceNet~\cite{kumar2019fisheyedistancenet} & 1024 x 320 & 0.109 & 0.788 & 4.669 & 0.185 & 0.889 & 0.964 & 0.982 \\
\textbf{SynDistNet}~\cite{kumar2021segdepth} & 1024 x 320   & \textbf{0.103} & \textbf{0.705} & \textbf{4.386} & \textbf{0.164} & \textbf{0.897} & \textbf{0.980} & \textbf{0.989} \\
\bottomrule
\end{tabular}
}
\vspace{-5pt}
\label{tab:recog_recon}
\end{table*}

Thus, we demonstrate the strong link between recognition and reconstruction. This idea is not particularly new. There was research investigating the potential of joint semantic labelling and depth as early as 2010 \cite{Liu-2010} (building upon even earlier work in geometric/semantic consistency in images \cite{Gould-2009}). 
However, it is fair to say that with the advent of neural networks in the last few years, the true potential of this research is beginning to come to fruition.

\subsection{Relocalization and Recognition}

\begin{figure}[ht]
  \captionsetup{singlelinecheck=false, font=small,  belowskip=-10pt}
  \centering
    \includegraphics[width=1.0\linewidth]{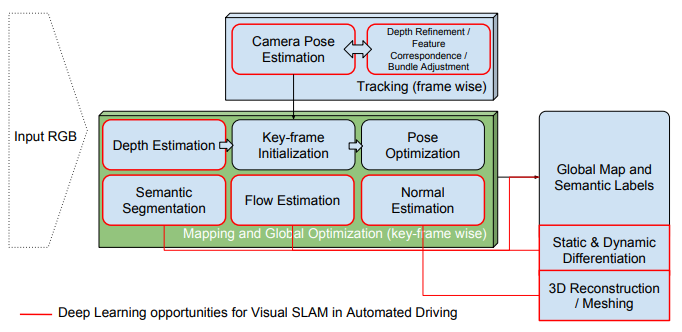}
    \caption{The Fundamental pipeline of Visual SLAM is composed of multiple geometric vision tasks including depth estimation, optical flow and pose estimation. Those tasks have well known solutions based on CNNs in their individual domain. In contrast, the overall Visual-SLAM is not dominated by Deep Learning. \cite{milz2018visual}}
    \label{fig:deepslam}
\end{figure}

Relocalization is the process of a vehicle recognizing a previously learned position or path, as discussed. However, in the real automotive world, many things can disturb this. For example, the scene can change due to movable objects - for example, parked vehicles can move between the time the scene is learned and when relocalization is requested. In such a case, semantic segmentation approaches, (e.g. \cite{siam2017deep, siam2018rtseg}), can be used to identify objects that may potentially move (vehicles, bicycles, pedestrians), and remove mapped features associated with such objects. Further opportunities exist for the support of traditional Visual-SLAM pipelines with deep learning techniques (Figure \ref{fig:deepslam}), as described in detail in \cite{milz2018visual}.

Place recognition in Visual-SLAM has a couple of applications. Firstly, it allows loop closure to correct for accumulated drift, and secondly it allows for building and maintaining maps from multiple passes through the same scene. Classical approaches using Bag of Words (e.g. \cite{Lopez2012}) proved reasonably successful, if perhaps lacking in terms of robustness. CNN-based approaches are proving to be more robust, with appearance-invariant approaches showing promising, if initial, results \cite{Garg2018}. The recognition of places when significant time has passed is an important topic. Table \ref{tab:relocrate} shows a small set of results for a Visual-SLAM pipeline, and demonstrates that errors increase significantly with a six-month time difference between training and relocalization.

\begin{table}
\centering
\caption{Quantitative results of our relocalization algorithm on selected WoodScape \cite{yogamani2019woodscape} dataset scenes. The time difference is the number of days between training and relocalization. The distance is the starting distance from the trained trajectory. The average offset is given in terms of position and angle.}
\begin{tabular}{|l|l|l|l|l|l|}
\hline
\multicolumn{2}{|c|}{\textbf{Scene}}                & \multicolumn{2}{c|}{\textbf{Difference}} & \multicolumn{2}{l|}{\textbf{Average Offset}} \\ \hline
\begin{tabular}[t]{@{}c@{}}Training\\Date\end{tabular}        & \begin{tabular}[t]{@{}c@{}}Replay\\Date\end{tabular}          & \begin{tabular}[t]{@{}c@{}}Time\\(days)\end{tabular}           & \begin{tabular}[t]{@{}c@{}}Dist.\\(m)\end{tabular}      & \begin{tabular}[t]{@{}c@{}}Pos.\\(m)\end{tabular}         & \begin{tabular}[t]{@{}c@{}}Angle\\(deg)\end{tabular}                          \\ \hline
20161208 & 20161208 & 0.003          & 4.723          & 0.468            & 4.704   \\ \hline
20161208 & 20161208 & 0.005          & 2.483          & 0.355            & 5.366   \\ \hline
20161208 & 20161208 & 0.006          & 2.692          & 0.3              & 5.149   \\ \hline
20161208 & 20170607 & 181.156        & 2.49           & 1.085            & 8.162   \\ \hline
20161208 & 20170607 & 181.155        & 0.066          & 0.903            & 9.498   \\ \hline
20161208 & 20170607 & 181.154        & 4.96           & 0.896            & 10.751  \\ \hline
\end{tabular}
\label{tab:relocrate}
\end{table}

Finally, view invariant localization can be considered. This is important when the camera viewpoint at the relocalization time is significantly different to the camera viewpoint at training, for example due to a rotation of the vehicle caused by approaching the trained trajectory at a large angle. Traditional Visual-SLAM methods based on feature descriptors fail, as the same surfaces of the landmarks may not even be visible. It has been shown that attaching semantic labels to scene land-marks (via bounding box classification) can significantly improve the performance of viewpoint invariance \cite{Li2019semanticslam}.

\subsection{Relocalization and Reconstruction}

This is perhaps the most straightforward synergy to discuss. Relocalization, and Visual-SLAM in general, can be considered as the storage of scene reconstruction (i.e., building a map) along with iterative refinement of said map through bundle adjustment (refer to Figure \ref{fig:reloc_pipeline}). In this way, reconstruction and visual odometry become a seed for the traditional Visual-SLAM approaches. There are direct methods that bypass this seeded approach, for example LSD-SLAM \cite{engel2014lsd} (and its Omnidirectional camera extension \cite{caruso2015large}), where photometric error is minimized as opposed to reprojection error. However, if one considers a time-slicing of a bundle-adjusted map, it can also be seen that Visual-SLAM can be used to refine the reconstruction (both scene structure and visual odometry). In addition, it is well known moving objects (as distinct from \textit{moveable} objects discussed in the previous section) can cause significant degradation in the performance of any Visual-SLAM pipeline \cite{Wangsiripitak2009}. Dynamic object detection (e.g. \cite{mariotti2020motion, siam2018modnet, yahiaoui2019fmodnet}) can therefore be used as an input into a Visual-SLAM pipeline to suppress outliers caused by said moving objects.

\begin{table*}[ht]
\centering
\caption{Features provided by current generation 4R architecture and comparison with previous and next generation. }
\begin{adjustbox}{width=2\columnwidth}
\begin{tabular}{|l|l|l|l|}
\hline
Module & Previous Gen & Current Gen 4R framework & Next Gen (Unified CNN \cite{ravikumar2021omnidet})   \\ \hline
Recognition & 
\begin{tabular}[c]{@{}l@{}}Pedestrian (PD) \\ Park-slot detection (PSD)\end{tabular} &
\begin{tabular}[c]{@{}l@{}}Bounding Box - PD, Cyclist, Vehicles\\ Segmentation - Road, curb, road markings\end{tabular} & \multirow{4}{*}{ \includegraphics[width=0.2\linewidth]{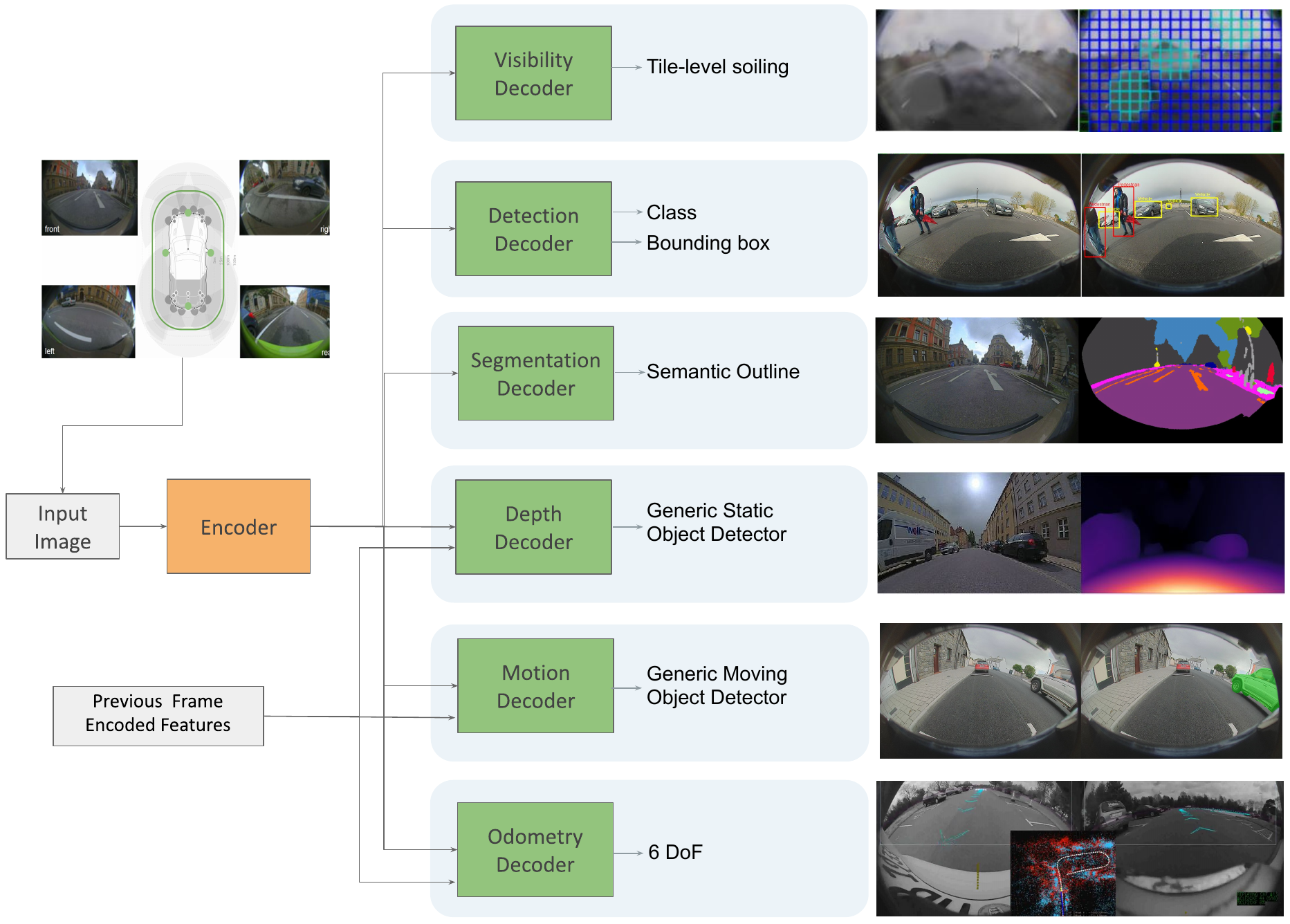} }\\ \cline{1-3}
Reconstruction & \begin{tabular}[c]{@{}l@{}}Sparse 3D Reconstruction\\ Sparse Flow\\ Visual Odometry(VO)\end{tabular} & \begin{tabular}[c]{@{}l@{}}Dense Depth\\ Dense Flow clustering\\ Multicam VO\end{tabular} &  \\ \cline{1-3}
Reorganization & - & \begin{tabular}[c]{@{}l@{}}Static Obj Fusion - Freespace, Curb\\ Dynamic Obj Fusion - Vehicles, PD, Cyclists\\ Lane Handler - Multicamera 3D lane fit\end{tabular} &  \\ \cline{1-3}
Relocalization & - & Sparse feature geometric map &  \\ \hline
\end{tabular}
\end{adjustbox}
\label{tab:compare-gen}
\end{table*}

\begin{figure*}[ht]
  \captionsetup{singlelinecheck=false, font=footnotesize, skip=2pt, belowskip=-8pt}
  \centering
    \includegraphics[width=\textwidth]{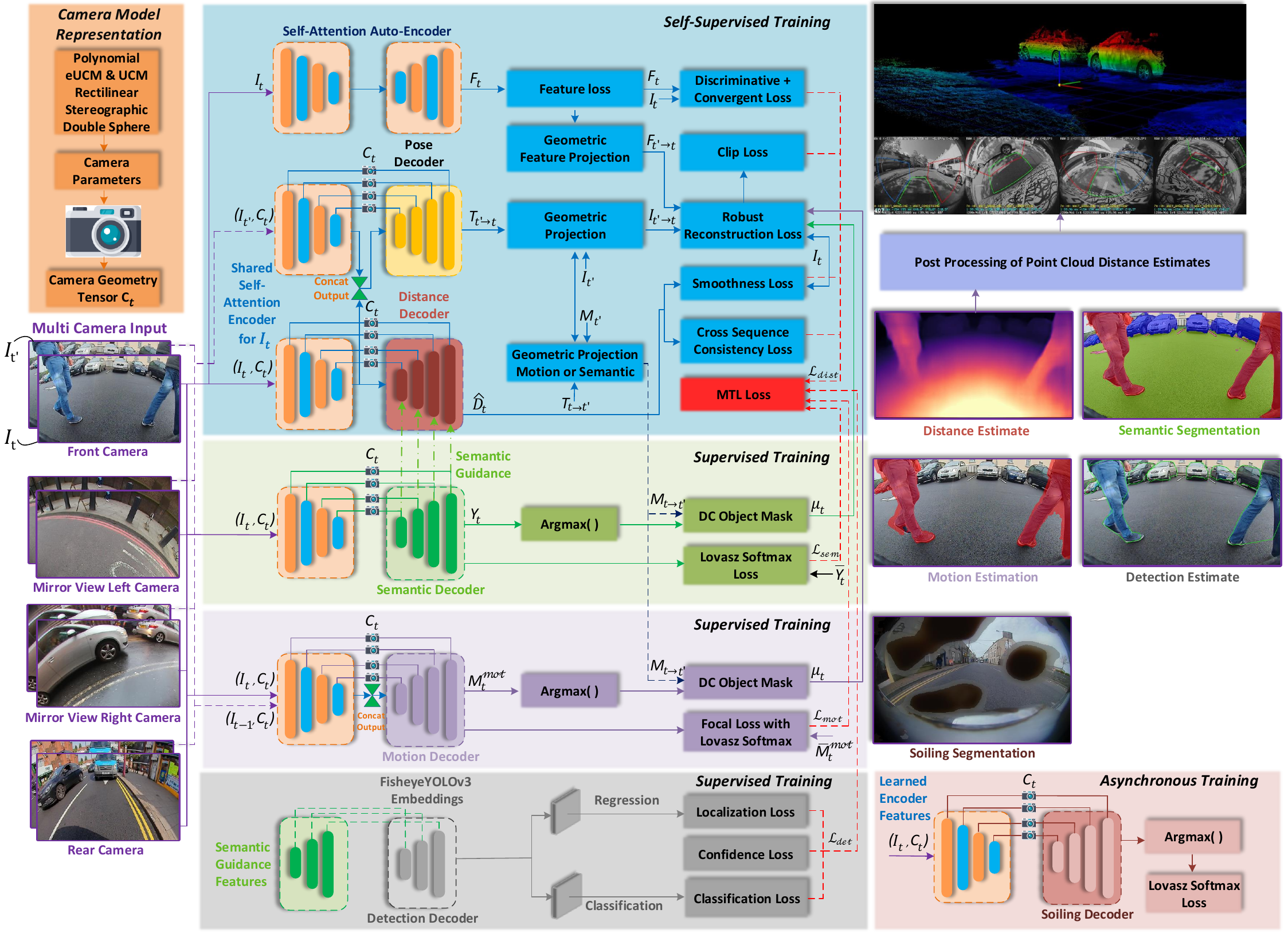}
    \caption{{Overview of our next generation unified multi-task visual perception framework.} Refer to our OmniDet paper \cite{ravikumar2021omnidet} for more details.
    }
    \label{fig:omnidet}
\end{figure*}

\subsection{Synergies in next generation}

Table \ref{tab:compare-gen} compares the current 4R architecture with previous and next generation architectures. Previous generation has simplistic features due to limited compute availability. For Recognition, it had only pedestrian and park-slot detection using classical machine learning. Reconstruction was performed using sparse optical flow in software without any hardware accelerators. There was no image level reorganization or relocalization. CNN models have progressed rapidly to provide state-of-the-art results for geometric tasks like reconstruction \cite{Qiao2021} and relocalization \cite{Oliveira2020}. For the next generation, a unified CNN model with high synergies would be the likely path. We have recently published an initial prototype Omnidet \cite{ravikumar2021omnidet} showing joint modelling of reconstruction and recognition.
Figure \ref{fig:omnidet} illustrates its high level architecture with cross links shown across the different tasks. 
In Table \ref{tab:results_next_gen}, an extract of the results from that work are presented. While this is not a complete implementation of what we would consider next generation, it does indicate that there is significant potential in jointly considering vision tasks.

While we believe that the proposal to use a complete CNN pipeline will offer optimal performance, we openly acknowledge that the approach has shortcomings. There are open challenges to improve the interpretability and trustworthiness of the perception model \cite{Schwalbe2020Safety}. Uncertainty estimation plays a critical role in providing the confidence of the prediction to effectively fuse with other sensor perception and it will also enable the determination of out of distribution input samples for safe handover to a manual driver. The ability to debug failures during development and deployment phase is another challenge which has recently received increased attention through explainable AI techniques \cite{zablocki2021explainability}. However, until interpretability and debuggability approaches for CNN-based processing is significantly more mature, we would propose that some level of redundancy is required by supplementing complete CNN-based approaches with classical computer vision and machine learning. We discuss this in more detail in the next section. In addition, it should be noted that some prominent scholars argue that improved performance is obtained by designing principled algorithms for the geometric estimations, and using deep neural networks for the extraction of robust visual features \cite{Sarlin2021}, where they argue that taking this approach results in a system capable of redeployment into new scenes without fine-tuning or retraining. That said, in future work, we plan to explore inclusion of relocalization and reorganization in neural network frameworks \cite{Cui2021}.

\begin{table*}
\centering
\caption{Extract of the results from our previous OmniDet work \cite{ravikumar2021omnidet}. It can be seen that jointly learning the tasks outperforms treating each task separately ($\downarrow$ means lower is better, $\uparrow$ means higher is better, PA denotes pixel accuracy). VarNorm task weighting is used. }
\begin{tabular}{ l c c c c c c c }
\hline
& \multicolumn{2}{c}{\begin{tabular}[t]{@{}c@{}}Distance\\Estimation\end{tabular}} 
& \multicolumn{2}{c}{\begin{tabular}[t]{@{}c@{}}Semantic\\Segmentation\end{tabular}}
& \multicolumn{2}{c}{\begin{tabular}[t]{@{}c@{}}Motion\\Segmentation\end{tabular}}
& \multicolumn{1}{c}{\begin{tabular}[t]{@{}c@{}}Object\\Detection\end{tabular}}\\ \hline
            & Sq. Rel $\downarrow$ & Abs Rel $\downarrow$ & mIoU $\uparrow$ & PA $\uparrow$ & mIoU $\uparrow$ & PA $\uparrow$ & mAP $\uparrow$ \\ \hline
Single Task & 0.060 & 0.304 & 72.5 & 94.8 & 68.1 & 94.1 & 63.5   \\
OmniDet     & 0.046 & 0.276 & 76.6 & 96.4 & 75.3 & 96.1 & 68.4   \\ \hline
\end{tabular}
\label{tab:results_next_gen}
\end{table*}

\subsection{Dual-sources of detection}

\begin{figure}[ht]
  \captionsetup{singlelinecheck=false, font=small,  belowskip=-10pt}
  \centering
    \includegraphics[width=1.0\linewidth]{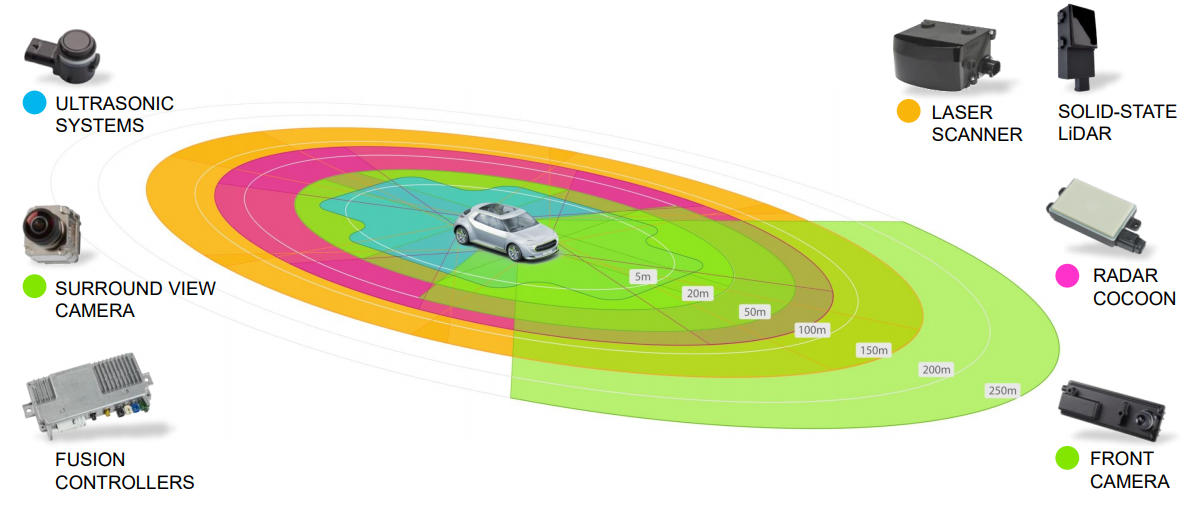}
    \caption{Perception cocoon of redundant vehicle sensors. Safety is achieved through at least one sensor being available if others become unavailable.}
    \label{fig:perceptioncocoon}
\end{figure}

We have thus far discussed possible synergies between Reconstruction, Recognition and Relocalization. There is, however, another overarching synergistic consideration: that of redundancy. In automated vehicles, redundancy plays a significant role in the safety of the application. When a system component fails, then another must be available to ensure that the vehicle remains in a safe state. For example, FuseModNet \cite{rashed2019fusemodnet} illustrates a synergistic fusion of cameras which provide dense information and lidar which performs well at low light. 
In terms of sensing, this would traditionally be achieved using multiple sensor types, such as computer vision systems, radar and laser scanner (Figure \ref{fig:perceptioncocoon}). For near-field sensing, an array of ultrasonic sensors is a mature low-cost sensor which provides robust safety around the vehicle \cite{popperli2019capsule}.

It is our contention that added safety is achieved through the parallel usage of different computer vision algorithm types. That is, a computer vision system architecture can be configured to maximize redundancy. This is particularly true as the sources are completely different types of processing – for example, statistical processing from the recognition pipeline and geometric processing from the reconstruction pipeline (Figure \ref{fig:4R_arch}). In addition, such processing will typically run-on different silicon components within an SoC. However, one must be aware that if you maximize the other synergies, the potential for redundancy is reduced. For example, if you use a CNN-based depth as a seed for a Visual-SLAM algorithm, you cannot claim the CNN as a redundancy for the Visual-SLAM, as Visual-SLAM is now dependent on the CNN processing. One must also be aware that the two processing elements will likely use the same video feed -- and so the safety of the camera itself and associated hardware/software, may also be a limiting factor. However, one should consider the potential for added safety in a system design following the 4R principles.

\section{Conclusions} \label{sec:conc}

In this paper, we provided a high-level survey of visual perception on surround-view cameras targeting commercial grade automated driving systems. We structure our survey into modular components namely Recognition, Reconstruction, Relocalization and Reorganization, jointly called 4Rs, and we argue that designing a vision architecture for vehicle automation along the lines of a 4R architecture can lead to system efficiencies. We discussed each component in detail and then we discussed how they are synergized to provide a more accurate system. We also provide a system and application context helping understand an industrial system. We have presented several architectures and frameworks, augmented with results predominantly from our previous publications, that support our argument in this direction.

The first three of the 4Rs (Recognition, Reconstruction, Relocalization) provide the means for the detection of objects and the extraction of their geometry and location in reference to the autonomous vehicle. However, that is not a complete description of the scene, and the fourth R (Reorganization) provides a higher-level scene understanding that can include the contextual spatial and temporal relationships between objects in the scene and the autonomous vehicle. Though massive advances have been made in the last decade in computer vision, we cannot yet claim to have achieved this complete scene understanding. It is likely that full vehicle autonomy will not be feasible until we have such a high level of visual reasoning deployed on vehicles. However, we propose that the 4R architecture can encapsulate, and provide a framework for, this level of vehicular cognition.

\section*{Acknowledgment}
We would like to thank our employer Valeo for encouraging advanced research. Many thanks to Edward Jones (NUI Galway) and Matthieu Cord (Sorbonne University and Valeo.ai) for providing a detailed review prior to submission. We would also like to thank our colleagues Fabian Burger, Nagarajan Balmukundan, Pantelis Ermilios and Nivedita Tripathi for supporting the paper.

\bibliographystyle{IEEEtran}
\bibliography{references}

\vfill

\end{document}